\documentclass[10pt]{article}
\PassOptionsToPackage{table}{xcolor}
\usepackage[accepted]{tmlr}

\usepackage{amsmath,amsfonts,bm}

\def\eqref#1{equation~\ref{#1}}

\def\1{\bm{1}}

\DeclareMathAlphabet{\mathsfit}{\encodingdefault}{\sfdefault}{m}{sl}
\SetMathAlphabet{\mathsfit}{bold}{\encodingdefault}{\sfdefault}{bx}{n}

\usepackage{hyperref}
\usepackage{url}

\usepackage{microtype}
\usepackage{graphicx}
\usepackage{subcaption}
\usepackage{booktabs}
\usepackage{multirow}
\usepackage{amssymb}
\usepackage{mathtools}
\usepackage{xcolor}
\usepackage{enumitem}
\usepackage{tabularx}
\usepackage{bbm}
\usepackage{tcolorbox}

\newcommand{\vocabV}{\mathcal{V}}
\newcommand{\realR}{\mathbb{R}}
\newcommand{\makered}[1]{\textcolor{red}{#1}}

\newcommand{\rev}[1]{#1}

\newcommand{\change}[1]{}
\newcommand{\wsinit}{\texttt{Warmstart-Init}}
\newcommand{\mttrain}{\texttt{Multitask-Train}}

\definecolor{indomain}{RGB}{198, 224, 180}
\definecolor{outdomain}{RGB}{252, 213, 180}
\definecolor{ashgrey}{rgb}{0.7, 0.75, 0.71}
\definecolor{lavendergray}{rgb}{0.77, 0.76, 0.82}
\definecolor{lemonchiffon}{rgb}{1.0, 0.98, 0.8}
\definecolor{amberdk}{RGB}{140, 90, 10}

\usepackage[capitalize,noabbrev]{cleveref}

\title{Training Continuous Chain of Thought Models: \\A Tale of Two Regimes}

\author{\name Varun Yerram \qquad He He \qquad Eunsol Choi \\
      \addr New York University \\
      \addr y.varun@nyu.edu
}

\def\month{Nov}
\def\year{2026}
\def\openreview{\url{https://openreview.net/forum?id=yz4LDTmIOD}}

\begin{document}

\maketitle

\begin{abstract}
Continuous Chain-of-Thought methods replace verbose reasoning traces with a short sequence of dense latent representations. Earlier continuous CoT methods indirectly supervise the latent representations such that their final states match that of verbose reasoning traces, requiring autoregressive, slow generation during training. We introduce C-MTP, a simpler, faster direct supervision approach that models each latent as an average of the embeddings in the CoT traces to be compressed. Our approach outperforms a prior direct supervision method that approximates the distribution of compressed tokens, and performs competitively with slower indirect supervision approaches in the existing evaluation setup with simplified CoT traces (less than 100 tokens). Lastly, we extend the evaluation of Continuous CoT methods to complex tasks with longer reasoning traces ($\ge$ a few hundred reasoning tokens). We find both direct and indirect supervision training methods perform poorly (roughly 60\% performance drop) in this setting, revealing the limitations of current continuous CoT methods. The code and checkpoints are released at \url{https://github.com/Varun221/cmtp_research}.
\end{abstract}

\begin{figure}[!ht]
    \centering
    \includegraphics[width=24em]{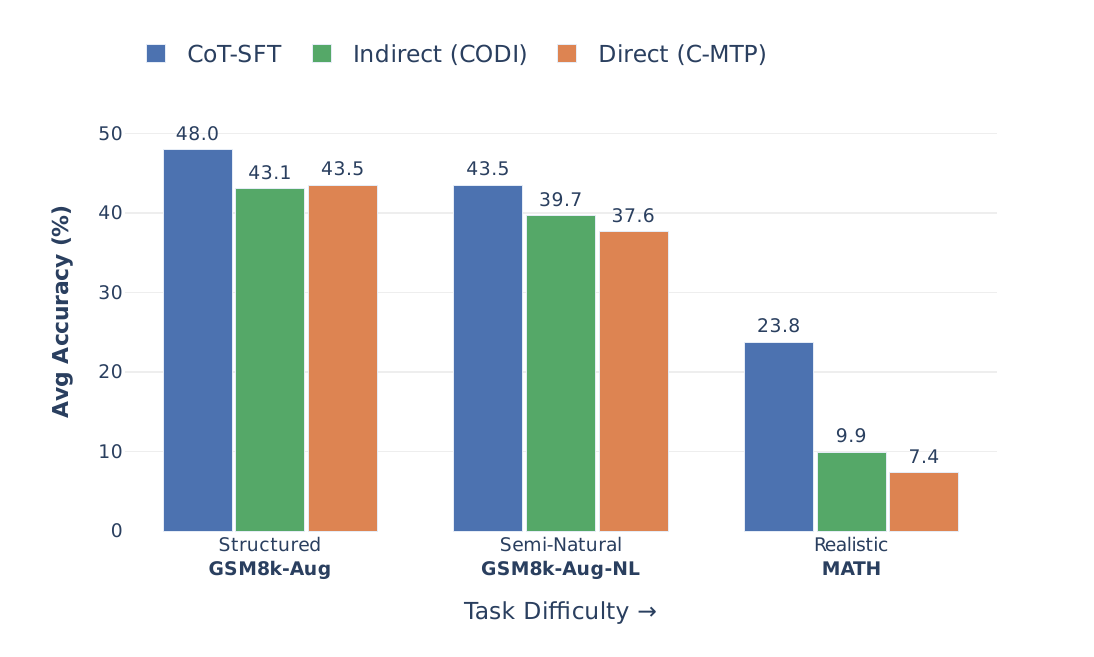}
    \caption{Head-to-Head comparison of standard CoT finetuning with two ContinuousCoT training methods: \textbf{Direct Supervision} (C-MTP, our proposed method) and \textbf{Indirect Supervision} (CODI) on three datasets of different CoT traces. \emph{Structured} consists of compact mathematical expressions of $\sim$25 tokens, \emph{Semi-Natural} contains sentence per-step explanations of $\sim$62 tokens while \emph{Realistic} captures traces generated from an instruct LLM ($\sim$350 tokens on average). Both methods remain competitive with CoT-SFT on structured/semi-natural traces but the best performing approach collapses to $\sim$40\% of its accuracy.
    }
    \label{fig:mainresults}
    \vspace{-10pt}
\end{figure}

\section{Introduction}
Recent LLMs achieve remarkable performance on complex reasoning tasks via generating a lengthy sequence of tokens that describes their reasoning process, often called chain-of-thought (CoT) traces \citep{DBLP:conf/nips/Wei0SBIXCLZ22,DBLP:conf/nips/KojimaGRMI22}, expanding the expressive power of LLMs \citep{DBLP:conf/iclr/MerrillS24}.
However, autoregressively generating long CoT traces increases the computational cost significantly.

 Continuous Chain of Thought (ContinuousCoT), pioneered by \citet{hao2025training}, allows an LLM to reason primarily in a short sequence of dense latent representations without discrete ``thinking'' tokens, significantly reducing reasoning length and, therefore, inference compute.
 Current approaches exceed the performance of discrete CoT on structured graph search tasks like ProsQA \citep{hao2025training} and ProntoQA \citep{DBLP:conf/iclr/Saparov023} and show competitive performance (about 10\% performance degradation) on math reasoning tasks. \citep{DBLP:journals/corr/abs-2110-14168, DBLP:journals/corr/abs-2405-14838}
 These methods \citep{DBLP:conf/emnlp/ShenYZHDH25,wei2026simcot} train a model to autoregressively predict intermediate latents and use a CoT teacher to supervise the final output. This allows the model to compress varying length CoT trajectories into a fixed number of latents.
 The sequential autoregressive passes required during training make it computationally expensive. As these methods utilize chain-of-thought traces \emph{indirectly}, we refer to them as using \textbf{indirect supervision} \rev{\change{12}where latents assume no particular structure and are supervised through a downstream objective}. These methods start from a pretrained discrete token LLM and finetune it to predict latents that are fed back into the model for a fixed number of steps.

Recent work, CoLaR \citep{tan2026think}, mitigates this by modeling latents as parameterized Gaussian distributions conditioned on token embeddings aggregated from a CoT teacher. This enables a parallelized training objective and brings the training distribution closer to pretraining. However, the model must predict a distribution over the latent space, which poses optimization challenges. Since this method utilizes CoT traces directly for training and prediction we term such methods as using \textbf{direct supervision} \rev{\change{12}where CoT traces \emph{define} the latent structure}. Unlike indirect supervision methods, direct supervision methods enable {parallel} training.

In this work, we introduce a new direct supervision approach: \textbf{C-MTP}, which models Multi-Token Prediction (MTP) to simulate latent reasoning. We formalize intermediate latents as aggregated token embeddings within a chain-of-thought, training the model to predict the tokens composing each latent rather than a continuous vector. This aligns with the pretraining objective while enabling fully parallel training via teacher forcing over CoT traces, eliminating the need for sequential autoregressive passes. Our evaluations show that C-MTP significantly outperforms existing direct supervision methods due to its simplicity and alignment with the pretraining objective. With small structured reasoning traces, C-MTP is faster, more sample-efficient and generalizes better to out-of-distribution problems (\cref{sec:structuredanal,sec:trainaug}) than indirect supervision methods.

On verbose semi-natural CoTs, we find that indirect supervision outperforms direct supervision, primarily because its recurrent training compresses long reasoning traces into few latents, while direct supervision is bounded by the original trace length and
suffers from error accumulation under teacher-forcing at inference. Finally, on moving to
\rev{\change{2}complex and longer} instruct-llm generated CoT traces, both regimes fall well short of CoT-SFT. This suggests that structured CoT evaluations, which dominate current benchmarks, overstate the practical utility of ContinuousCoT methods.

We state our contributions as follows:

\begin{itemize}[nosep]
    \item We introduce \textbf{C-MTP}, a novel direct supervision framework that uses Multi-Token Prediction to simulate continuous latent reasoning.
    \item On structured reasoning tasks, C-MTP outperforms existing direct supervision methods and matches or exceeds indirect supervision methods while being faster, more sample-efficient, and generalizing better to out-of-distribution problems.
    \item We conduct a systematic empirical comparison across structured, semi-natural and realistic CoT trace regimes, showing that indirect supervision overtakes direct on verbose semi-natural traces through its compression objective, and that both regimes fall short of CoT-SFT on realistic (\rev{\change{2}i.e. complex and longer}) traces, exposing the limitations of ContinuousCoT approaches.

\end{itemize}

\section{Related Work}
\label{app:relatedworks}

Chain-of-Thought (CoT) \citep{DBLP:conf/nips/Wei0SBIXCLZ22,DBLP:conf/nips/KojimaGRMI22} has found a wide range of applications. It improves self-consistency \citep{DBLP:conf/iclr/0002WSLCNCZ23}, simplifies complex problem solving through decomposition \citep{DBLP:conf/iclr/ZhouSHWS0SCBLC23}, facilitates planning \citep{wang-etal-2023-plan}, automates demonstration generation \citep{DBLP:conf/iclr/0001Z0S23}, and enhances supervised finetuning \citep{DBLP:conf/iclr/WeiBZGYLDDL22,DBLP:journals/jmlr/ChungHLZTFL00BW24,DBLP:conf/nips/ZelikmanWMG22,DBLP:conf/acl/HsiehLYNFRKLP23}. These advancements have culminated in recent ``thinking'' models \citep{guo2025deepseek,openai2026openaio1card}.

ContinuousCoT, or latent reasoning, encompasses diverse approaches. COCONUT \citep{hao2025training} pioneered autoregressive generation with latents, inspiring subsequent research. Formal analyses \citep{DBLP:journals/corr/abs-2505-12514,DBLP:journals/corr/abs-2602-01148,DBLP:journals/corr/abs-2509-25239} demonstrate its theoretical advantages over explicit CoT in specific scenarios. In this context, \emph{latents} can encode multiple solution trajectories \citep{DBLP:journals/corr/abs-2505-23648, DBLP:journals/corr/abs-2505-15778, wu2025llmssinglethreadedreasonersdemystifying, DBLP:journals/corr/abs-2601-08808, butt2025softtokenshardtruths}, provide additional compute via filler tokens \citep{DBLP:journals/corr/abs-2404-15758, DBLP:conf/iclr/GoyalJRMKN24}, or use recurrent looping \citep{DBLP:conf/iclr/SaunshiDLKR25, DBLP:journals/corr/abs-2511-08577, zeng2026ponderlmpretraininglanguagemodels, DBLP:journals/corr/abs-2510-25741, DBLP:journals/corr/abs-2502-05171}.

In this work, we propose C-MTP, a \rev{\change{16}representative} direct supervision approach \citep{tan2026think, DBLP:journals/corr/abs-2412-13171} that outperforms the current best method in the regime.  Over various trace lengths and problem difficulties we compare this regime with approaches in indirect supervision  \citep{DBLP:conf/emnlp/ShenYZHDH25, wei2026simcot}. More recent works such as \citet{DBLP:journals/corr/abs-2606-31779} connect indirect supervision to looped LLMs by initializing latent slots and iteratively refining them. Concurrently, the field is advancing novel training architectures \citep{wu-etal-2025-parallel, shao2025continuousautoregressivelanguagemodels} and hybrid models that adaptively switch between latent reasoning and token-based generation \citep{shi2026swireasoningswitchthinkinglatentexplicit, yue2025hybridlatentreasoningreinforcement, xu2026thinkrouterefficientreasoningrouting, zhu2026lightthinkerreasoningcompressionmemory, zhang2025lightthinkerthinkingstepbystepcompression}. Our work underscores the need for this adaptivity. \rev{\change{17}Latent reasoning can also be learned through variational objectives \citep{DBLP:journals/corr/abs-2509-25020, Wang2026ReGuLaR, kang2026ladir}, a promising orthogonal direction.}

Finally, while Multi-Token Prediction (MTP) improves pretraining \citep{deepseekai2025deepseekv3technicalreport, DBLP:conf/icml/GloeckleIRLS24}, accelerates speculative decoding \citep{DBLP:conf/icml/CaiLGPLCD24}, and can be learned via self-distillation \citep{DBLP:journals/corr/abs-2603-23911,kirchenbauer2026multitoken}, it remains largely unexplored in the context of latent reasoning.

\section{Methodology}

\subsection{Notation and Problem Setup}

Consider an LLM $M_\theta$. Let $\vocabV$ be its vocabulary and $E \in \realR^{|\vocabV| \times d}$ be the embedding matrix. Given an input prompt $Q = (q_1, q_2, \ldots, q_m) \in \vocabV^m$
we denote the embedded prompt as $E(Q) = (e(q_1), e(q_2), \ldots, e(q_m))$, where $e(q_i) \in \realR^d$. Given a prompt, we can obtain the final layer hidden state at the final position from a student model $M_\theta$ with a function denoted by $H_\theta(.) \in \realR^d$.

Let's consider that we have access to $K$ reasoning steps for the given prompt. Denote them as $y := (y_1, y_2, \ldots, y_K)$. Each $y_i = (y_{i1}, y_{i2}, \ldots, y_{iL_i}) \in \vocabV^{L_i}$ and $L_i$ is the token length of the $i$-th textual reasoning step. Finally, we denote an answer as $a = (a_1, a_2, \ldots, a_n) \in \vocabV^n$.

~\\
Standard CoT supervised finetuning optimizes the following objective:
\begin{align}
    L_{CoT} = - \log P_\theta(a \mid Q, y_1, y_2, \ldots, y_K)
    \label{eqn:cotloss}
\end{align}
Where $P_\theta(.)$ is the probability distribution defined by the student model $M_\theta$. (Through $H_\theta(.)$ and a softmax over the vocabulary). Continuous CoT aims to replace textual reasoning tokens with a sequence of dense latents $z := (z_1, z_2, \ldots, z_K)$\footnote{Note that the number of latents need not equal reasoning steps of a problem. Here we consider equal case for simplicity.} obtained by
\begin{align}
    z_k = H_\theta(I^{(k-1)}) \;\;\;\; I^{(k)} = I^{(k-1)} \oplus z_k
\end{align}
Where, $I^{(0)} = E(Q)$ is the embedded input prompt and each $z_i \in \realR^d$. $\oplus$ denotes concatenation along the sequence axis. The Continuous Chain of Thought is therefore defined as the sequence of continuous vectors $z := z_1, z_2, \ldots, z_K$. The primary goal of all ContinuousCoT methods is to design an effective training recipe to learn $z$ such that the final answer $a$ can be generated conditioned only on $z$ and $Q$.

\begin{table}[t]
    \centering
    \small
    \setlength{\tabcolsep}{2pt}
    \caption{
    \textbf{Comparison of indirect and direct supervision regimes for ContinuousCoT training.}
Indirect methods reuse the model's own final-layer activation via recurrence, producing continuous latents. Direct methods instead map predictions through the embedding table, constraining outputs to the token space. For indirect methods, the reasoning length (number of recurrent latent steps) is a fixed hyperparameter set before training, whereas direct methods operate over variable-length CoT traces via teacher forcing. \rev{\change{12}We note that all properties follow from whether CoT traces define the latent \emph{structure} (\cref{app:definitionexplain}), and describe the regimes as instantiated by the methods in our evaluated setting (\cref{sec:limitations}). }}
    \begin{tabular}{l cc}
        \toprule
         & \textbf{Indirect} & \textbf{Direct} \\
         & \textbf{Supervision} & \textbf{Supervision} \\
         & (e.g., CODI, Sim-CoT, & (e.g., CoLaR,\\
         & COCONUT) & C-MTP)\\
        \midrule
        \multirow{2}{*}{Input} & Final Layer & Aggregate of \\
         & Activation & Token Embeddings \\
        \midrule
        Output & Continuous & Discrete \\
        \midrule
        Training & Slow & Fast \\
        Efficiency & ($>1$ Forward passes) & ($1$ Forward pass) \\
        \midrule
        Reasoning & \multirow{2}{*}{Fixed} & \multirow{2}{*}{Variable} \\
        Length & & \\
        \midrule
        Training Method & Recurrence & Teacher Forcing \\
        \bottomrule
    \end{tabular}
    \label{tab:supcomparison}
\end{table}
\subsection{Prior Work: Indirect Supervision Methods}
\label{sec:priorworkindir}

Coconut \citep{hao2025training} uses a curriculum-based strategy where at each iteration, one of the textual reasoning steps $y_i$ is replaced with the corresponding latent $z_i$ starting from $i=0$. The following objective is optimized at each step $i$.
\begin{align}
    L^{\text{curriculum}}_{i} = - \log P_\theta(a \mid Q, z_1, \ldots, z_i, y_{i+1}, \ldots, y_K)
\end{align}
This progressively teaches $M_\theta$ to generate semantically useful latents and eventually answer based on these latents alone. CODI \citep{DBLP:conf/emnlp/ShenYZHDH25} improves training efficiency by replacing curriculum with distillation.
The student, $M_\theta$ autoregressively generates $K$ latents during training. Using these latents, it is trained to generate the answer.
\begin{align}
    L_{Student} = - \log P_\theta(a \mid Q, z_1, z_2, \ldots, z_K)
\end{align}
The teacher $M_\phi$ (warm started with $\theta$) is trained using standard CoT traces using \cref{eqn:cotloss}\footnote{\citet{DBLP:conf/emnlp/ShenYZHDH25} uses multi-task training where student and teacher share the same weights. We see the effect of this in more detail in Appendix \ref{app:multitask}}.  The intermediate layer activations of $M_\phi$ are distilled to $M_\theta$ to guide student training substantially improving training efficiency and downstream performance.
\begin{align}
    L_{KD} = \frac{1}{D} \sum_{l=1}^{D} | sg[h_{\phi}^l] - h_{\theta}^l  |
    \label{eqn:distill}
\end{align}
Where $D$ is the number of layers in the model, $sg$ denotes stop-gradient, and $h^l$ denotes the $l$-th layer activation of the token before the start of answer generation, indexed by the corresponding model ($\phi$ for teacher, $\theta$ for student).
Recently, \citet{wei2026simcot} argued that when $K$ is increased to larger values, the latents degenerate to similar representations. To stabilize this effect they introduce step-level supervision using intermediate reasoning steps $y$.  A decoder LLM $N_\psi$ is trained to autoregressively generate the $k$-th reasoning step conditioned only on the latent $z_k$.
The authors show that adding this step-level supervision to CODI or Coconut improves performance significantly on downstream tasks.

These works share three characteristics: First, the number of latent steps is fixed prior to training, resulting in fixed-length reasoning steps. Second, the sequential, autoregressive generation of latents during training imposes significant overhead. Finally, these methods leverage textual reasoning steps \textbf{indirectly} for supervision; either through  teacher distillation or by using latents to decode them back.

\subsection{Prior Work: Direct Supervision Methods}
\label{sec:priorworkdir}

CoLaR \citep{tan2026think} proposed a method to utilize reasoning steps directly by compressing embeddings of every $s$ consecutive tokens in reasoning trace $y$ into a latent $z_i$. During training, one token is sampled from each $s$-token group to form a subsampled reasoning trace. Model $M_\theta$, is then trained via teacher forcing to predict this sampled trace and the final answer. This approach trains the model's predictions to approximate a multi-modal distribution representing all potential tokens within each compressed group. The aggregated embeddings are modeled as parameterized Gaussians, and a dedicated latent head is trained to predict the mean and standard deviation for each step. During inference, they use the reparameterization trick to sample from this distribution, passing the resulting latent vector to the next autoregressive step.

\subsection{C-MTP: Continuous Chain of Thought via Multi-Token Prediction}
\label{sec:cmtp-desc}

Similar to CoLaR, we view intermediate latents as aggregated token embeddings. However instead of modeling them as Gaussians we provide a simpler recipe using Multi-Token Prediction \citep{DBLP:conf/icml/GloeckleIRLS24}.\footnote{We describe our exact architecture in \cref{app:mtp-arch}}

During training, we set a desired compression factor $s$, the number of tokens per span.
Given the reasoning trace $y = (c_1, \ldots, c_L)$, we partition it into
$r = \lceil L/s \rceil$ non-overlapping spans of length $s$:
\begin{align}
    C_i = (c_{(i-1)s+1}, \ldots, c_{is}), \quad i = 1, \ldots, r
\end{align}
Each span is compressed into a single latent by averaging its token embeddings\footnote{\citet{tan2026think} divides by $\sqrt{s}$ instead of $s$. We did not find any difference in our experiments and hence use simple average for simplicity.}:
\begin{align}
    z_i = f(C_i) = \frac{1}{s} \sum_{j=1}^{s} E\!\left(c_{(i-1)s+j}\right)
    \label{eqn:embagg}
\end{align}
This yields a sequence of $r$ latents $z = (z_1, \ldots, z_r)$ that replaces
the full reasoning trace $y$ during the forward pass. At each token position, we want to model $P_{\theta}(z_i \mid Q, z_1 \ldots z_{i-1})$. Using \cref{eqn:embagg}, we cast the problem of next latent prediction as predicting the tokens composing the latent. We train the student, $M_\theta$ to minimize
\begin{align}
    L_{multi} = - \sum_{i=1}^{r} \sum_{t \in C_{i}} \log P_\theta(t \mid E(Q), z_1, \ldots, z_{i-1})
    \label{eqn:multiloss}
\end{align}
The answer supervision stays unchanged
\begin{align}
    L_{ans} = - \sum_{j=1}^{n} \log P_\theta(a_j \mid E(Q), z_{1:r} )
    &&
    L_{MTP} = L_{multi} + L_{ans}
    \label{eqn:anssup}
\end{align}
To guide training, we also employ a CoT Teacher for distillation using $L_{KD}$ (Defined in ~\cref{eqn:distill}). Overall, we employ two training configurations in this work: \textbf{\wsinit} where we first train a CoT teacher using $L_{CoT}$, use it to initialize the student, and then train the student using $L_{MTP} + L_{KD}$; and \textbf{\mttrain} where we train the same backbone to optimize both $L_{CoT}$ and $L_{MTP} + L_{KD}$ simultaneously. We found that \wsinit{} performs better in out-of-domain tasks and keep it for main experiments, we use \mttrain{} for fair comparison with CODI in relevant experiments. We present ablations for \wsinit{} \& \mttrain{} and the effect of $L_{KD}$ in \cref{sec:ablations}, showing that C-MTP is effective even without distillation from a teacher; we retain $L_{KD}$ in the main experiments to provide the strongest possible performance for C-MTP and to keep the training setup consistent with CODI which also uses distillation.

During inference, the trained $M_\theta$ first goes through a CoT phase and then answer phase.
In the CoT phase, at each step $i$ the model samples $s$ tokens from
$P_\theta(\cdot \mid E(Q), z_1, \ldots, z_{i-1})$, aggregates their embeddings to
form the next latent $z_i$ via \cref{eqn:embagg}, and appends it to the latent
sequence. This repeats until the end-of-thinking token is among the sampled
tokens. In the answer phase, the model switches to standard autoregressive
next-token prediction conditioned on $E(Q)$ and the accumulated latents
$z_{1:r}$, continuing until the end-of-sequence token is produced.

\begin{table*}[h]
    \caption{\textbf{GSM8k-Aug results on LLaMA-3.2-1B-Instruct.} Accuracy is averaged over 5 runs (run-to-run deviation $<$1\%). \#L denotes the mean chain-of-thought length (in steps) with standard deviation, computed over the full evaluation set.}
    \centering
    \setlength{\tabcolsep}{2pt}
    \small
    \begin{tabular}{ll cc cc cc cc}
        \toprule
        Training Method & & \multicolumn{2}{c}{GSM8k-Test} & \multicolumn{2}{c}{GSM-Hard} & \multicolumn{2}{c}{SVAMP} & \multicolumn{2}{c}{MultiArith} \\
        \cmidrule(lr){3-4} \cmidrule(lr){5-6} \cmidrule(lr){7-8} \cmidrule(lr){9-10}
        & & Acc. ($\uparrow$) & \# L ($\downarrow$) & Acc. ($\uparrow$) & \# L ($\downarrow$) & Acc. ($\uparrow$) & \# L ($\downarrow$) & Acc. ($\uparrow$) & \# L ($\downarrow$) \\
        \midrule
        \rowcolor{gray!15} Standard & CoT & 61.94 & $27.17_{\pm 11.24}$ & 14.22 & $34.43_{\pm 15.11}$ & 64.96 & $13.15_{\pm 5.30}$ & 98.89 & $15.46_{\pm 1.96}$ \\
        \midrule
        Indirect & Coconut\footnotemark & 45.30 & 6.00 & 9.9 & 6.00 & 48.80 & 6.00 & 90.10 & 6.00 \\
        Supervision & CODI & 53.03 & 6.00 & 12.40 & 6.00 & 60.26 & 6.00 & \textbf{99.11} & 6.00 \\
        & Sim-CoT & \textbf{53.95} & 6.00 & 12.45 & \textbf{6.00} & 58.62 & 6.00 & 97.78 & 6.00 \\
        \midrule
        Direct & CoLaR-2 & 41.23 & $12.99_{\pm 5.53}$ & 8.87 & $16.30_{\pm 8.30}$ & 52.46 & $6.31_{\pm 2.66}$ & 90.56 & $7.39_{\pm 1.43}$ \\
        Supervision & CoLaR-3 & 32.65 & $9.01_{\pm 3.75}$ & 6.79 & $10.96_{\pm 6.30}$ & 48.36 & $4.49_{\pm 1.91}$ & 87.33 & $5.33_{\pm 1.93}$ \\
        & CoLaR-4 & 29.60 & $6.80_{\pm 2.71}$ & 6.60 & $7.66_{\pm 2.97}$ & 46.82 & $3.47_{\pm 1.48}$ & 88.11 & $4.13_{\pm 0.67}$ \\
        \cmidrule(lr){2-10}
        & C-MTP-2 & 53.30 & $12.07_{\pm 5.53}$ & \textbf{12.59} & $14.15_{\pm 6.44}$ & \textbf{62.10} & $5.14_{\pm 2.58}$ & 95.33 & $6.29_{\pm 1.20}$ \\
        & C-MTP-3 & 46.50 & $7.94_{\pm 3.69}$ & 10.58 & $9.02_{\pm 4.54}$ & 56.08 & $3.43_{\pm 1.66}$ & 97.33 & $4.16_{\pm 0.67}$ \\
        & C-MTP-4 & 40.70 & \bm{$5.84_{\pm 2.71}$} & 9.29 & $6.85_{\pm 4.34}$ & 50.98 & \bm{$2.39_{\pm 1.47}$} & 94.67 & \bm{$3.12_{\pm 0.56}$} \\
        \bottomrule
    \end{tabular}
    \label{tab:trainaugllama}
\end{table*}
\footnotetext{Values taken from \citet{DBLP:conf/emnlp/ShenYZHDH25}}

\section{Experimental Results and Analysis}

In this section we discuss the setup, training details, experiments and detailed analysis of the proposed architecture.

\subsection{Setup}

\textbf{Datasets and Tasks.}  We primarily use \textbf{GSM8k-Aug} \citep{DBLP:journals/corr/abs-2405-14838}, an augmented version of Grade-School level Math reasoning dataset GSM8k \citep{DBLP:journals/corr/abs-2110-14168}. It comprises $\sim$385k training samples of which we reserve 1\% for validation and use the rest for training. It has proven effective for training ContinuousCoT methods as it does not contain any natural language but only structured mathematical expressions leading to the final answer such as $\ll 18.5 - 8.5 = 10 \gg  \ll 20+15+30=65  \gg$. We also use \textbf{GSM8k-Aug-NL}, a version that preserves natural language explanations. As each sentence perfectly corresponds to one reasoning step in structured trace, we term this dataset as Semi-Natural.
For experiments with realistic CoT traces, we use challenging \textbf{MATH} \citep{hendrycksmath2021} problems and generate CoT traces from the target model being trained. For evaluation, we use the test splits of GSM8k to evaluate in-domain performance, and \textbf{GSM-Hard}~\citep{pmlr-v202-gao23f}, \textbf{SVAMP}~\citep{patel-etal-2021-nlp} and \textbf{MultiArith}~\citep{roy-roth-2015-solving} to evaluate out-of-domain generalization on math tasks. We also use \textbf{MATH} test set for more complex natural language reasoning evaluation. See \cref{app:exampletraces} for example traces.

\textbf{Baselines.}
We consider the following baselines: (1) \textbf{CoT} (Chain-of-Thought) finetuning, where the model is trained to generate CoT tokens followed by the final answer. We compare with two indirect supervision methods: (2) \textbf{CODI} \citep{DBLP:conf/emnlp/ShenYZHDH25} and (3) \textbf{Sim-CoT} \citep{wei2026simcot}. For direct supervision we compare with (4) \textbf{CoLaR} \citep{tan2026think}. These methods are discussed extensively in \cref{sec:priorworkindir,sec:priorworkdir}. We use their official implementations and training scripts for our experiments. For CoLaR, we compare with three compression factors $s \in \{2, 3, 4\}$ which approximately correspond to the ratio of original chain of thought and generation length of the trained model.

\textbf{Training and Implementation Details.}
We experiment with two backbones: \textbf{LLaMA-3.2-1B-Instruct} \citep{grattafiori2024llama3herdmodels} and \textbf{Qwen2.5-1.5B-Instruct} \citep{qwen2025qwen25technicalreport}. Following prior works, all experiments use a trainable LoRA module \citep{hu2021loralowrankadaptationlarge} and are initialized with a trained CoT checkpoint. Full training and implementation details are provided in \cref{app:expdetails}. For C-MTP, we use \wsinit{} for the main experiments, and \mttrain{} for analysis in \cref{sec:structuredanal} to ensure a fair comparison with CODI. We present a detailed comparison of the two training configurations along with analysis of effect of $L_{KD}$ in \cref{sec:ablations}.

\subsection{Training on Structured CoT: GSM8k-Aug}
\label{sec:trainaug}

\cref{tab:trainaugllama,tab:trainaugqwen} compare C-MTP against the baselines from \cref{sec:priorworkindir,sec:priorworkdir} on GSM8k-Aug, reporting downstream accuracy and chain-of-thought length. \textbf{C-MTP outperforms CoLaR} across all compression ratios: on LLaMA-3.2-1B-Instruct, C-MTP at 4$\times$ compression matches CoLaR at 2$\times$, while converging in $\sim$5 epochs versus CoLaR's $\sim$50, highlighting C-MTP's stronger alignment with the pretraining objective.
Across models, neither approach consistently dominates. On LLaMA, indirect supervision outperforms direct on in-distribution data but underperforms on out-of-distribution; on Qwen (see \cref{app:structuredqwen}), this pattern reverses. The lack of a consistent winner across both models and distribution settings suggests the two approaches are broadly comparable, with relative performance being model-dependent. The following sections explore potential reasons why direct supervision can be preferable.

\subsubsection{Simplicity, Efficiency, and Generalization}
\label{sec:structuredanal}

The simplicity of the proposed direct supervision approach, C-MTP, yields practical advantages such as faster training, stronger sample efficiency, and better generalization to unseen problem structures. We analyze each of these dimensions below, comparing C-MTP against CODI under controlled settings. \rev{\change{10}To ensure fair comparisons, we use \mttrain{} for all results in this section}.

\textbf{Fixed Wallclock Time Comparison.} Based on our analysis in \cref{sec:timing_analysis}, we estimate that a single C-MTP training step is approximately $2\times$ faster than CODI on LLaMA-3.2-1B-Instruct, allowing C-MTP to process roughly twice as many samples within the same wallclock budget. \rev{\change{9}We note that this $2\times$ factor is an estimate of the average speedup across the sequence lengths and batch sizes that C-MTP can encounter against CODI. We present a detailed account of speedup estimation in \cref{sec:timing_analysis}. As actual training wallclock time estimates can be noisy depending on hardware and framework,} we allocate C-MTP's speedup toward doubling the training samples while holding epochs constant (details in \cref{app:fixwallclock}). We observe that C-MTP-4 consistently outperforms CODI across all budget levels in \cref{tab:fixedwallclock} (Appendix), with gains on GSM8k-Test particularly pronounced.

\textbf{Sample Efficiency.} To understand how the two training paradigms compare under varying amounts of supervision, we train both methods on randomly sampled subsets of GSM8k-Aug (12k, 25k, 50k, 100k and 200k samples) and evaluate on downstream tasks. For each subset size, we sweep over a predefined set of hyperparameters and select the best-performing configuration by validation loss. \cref{fig:sampleff} (Appendix) shows accuracy as a function of training set size. In the low-data regime (up to $\sim$100k samples), C-MTP consistently outperforms CODI, with the gap most pronounced at 12k samples. On in-distribution tasks (GSM8k-Test, GSM-Hard), C-MTP's advantage is particularly clear, whereas on out-of-distribution tasks (SVAMP, MultiArith) the gap is smaller and more variable.

\textbf{Generalization to Problem Structures.} GSM8k-Aug \citep{DBLP:journals/corr/abs-2405-14838} was constructed by prompting GPT-4 to generate problems similar to GSM8k, thereby expanding the dataset $\sim50\times$ via prompting and heuristic filtering. This process results in  substantial structural redundancy: our analysis shows that over 50\% of all training samples reduce to only 100 distinct problem structures. We present more analysis in \cref{app:gsmauganal}. To evaluate out-of-domain generalization to problem structures not seen during training, we partition the data by held-out problem structures for each reasoning length $n \in \{2,3,4\}$.

\cref{tab:gridsall} (Appendix) reports in-domain and out-of-domain accuracy for C-MTP and CODI across reasoning lengths $n \in \{2, 3, 4\}$. C-MTP generalizes better to held-out problem structures in most settings. The gap is particularly clear at $n=3$ and $n=4$, where CODI's out-of-domain accuracy drops substantially (e.g., 27.10 and 13.83 on $n=3$ and $n=4$ respectively) while C-MTP retains stronger performance (42.86 and 24.04). In-domain, the two methods are more competitive, with CODI performing slightly better at $n=2$ while C-MTP leads at longer reasoning chains. Overall, these results suggest that the sequential training of indirect supervision methods leads to overfitting to the specific problem structures seen during training. Full details on partitioning and training are provided in \cref{app:genproblems}.

\begin{figure}[h]
    \centering
    \captionsetup[subfigure]{font=small}
    \begin{subfigure}[t]{0.32\textwidth}
        \centering
        \caption{Self-consistency by position\\ (Structured GSM8k-test).}
        \label{fig:experroraccum}
        \includegraphics[width=\linewidth]{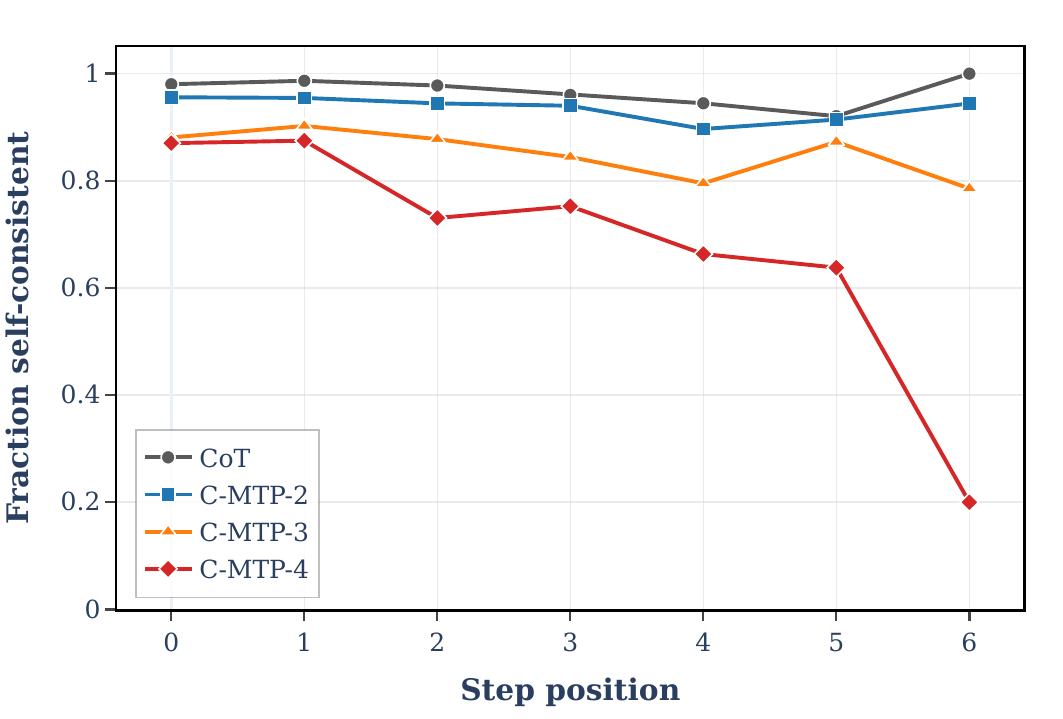}
    \end{subfigure}\hfill
    \begin{subfigure}[t]{0.32\textwidth}
        \centering
        \caption{Gold-prefill sweep\\ (Structured GSM8k-test).}
        \label{fig:gsmstructgsm_goldfill}
        \includegraphics[width=\linewidth]{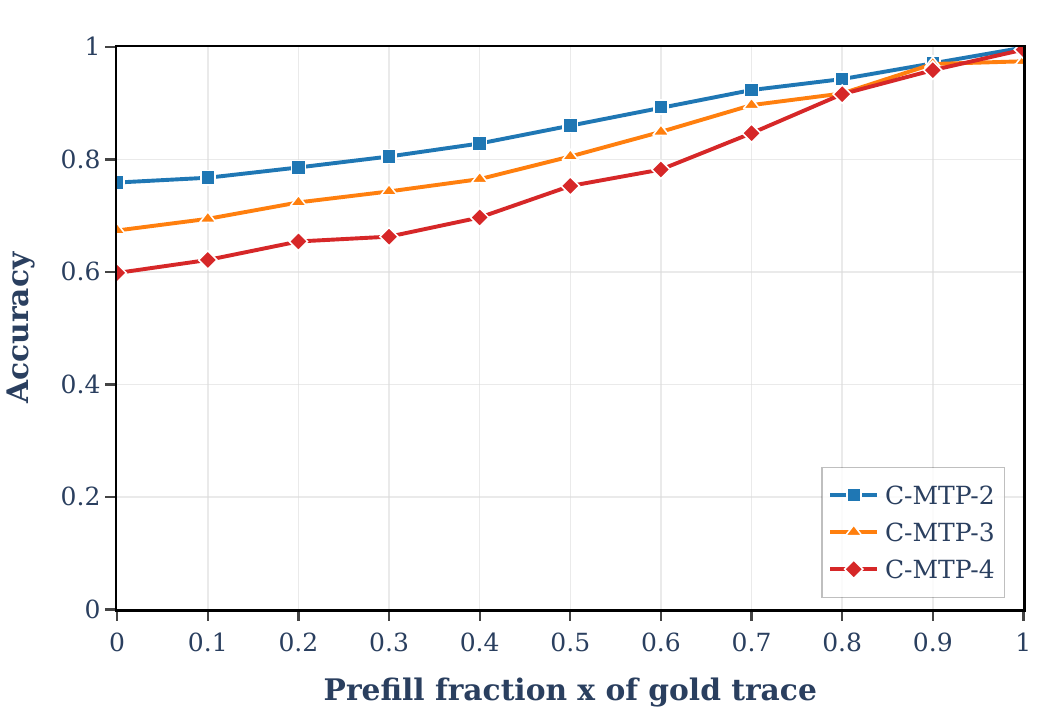}
    \end{subfigure}\hfill
    \begin{subfigure}[t]{0.32\textwidth}
        \centering
        \caption{Gold-prefill sweep\\ (Natural Language GSM8k-test).}
        \label{fig:gsmnl_goldfill}
        \includegraphics[width=\linewidth]{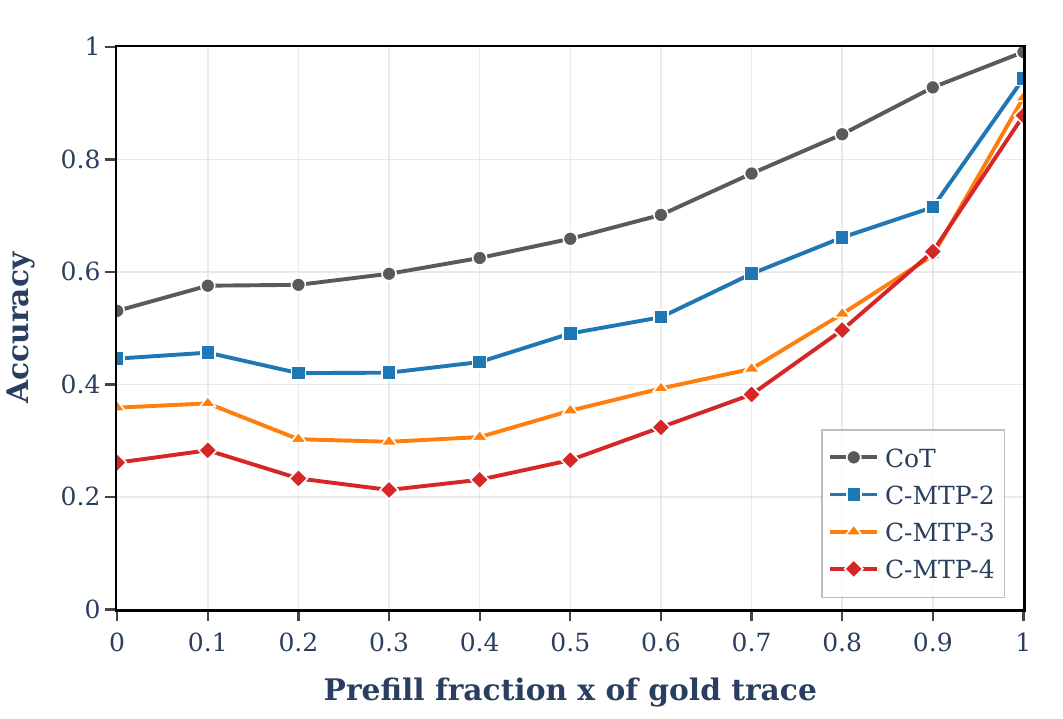}
    \end{subfigure}
    \caption{\rev{\change{4} \textbf{Prediction \& Accumulation of errors in spans, rather than token aggregation, accounts for most of C-MTP's degradation.} (a) Fraction of self-consistent generated expressions by position in the trace (GSM8k-Aug, evaluated on GSM8k-test): C-MTP starts near the CoT level early in the trace and falls off toward the end, with the drop growing at higher compression $s$. (b, c) Gold-trace prefill: we feed a fraction $x$ of the oracle gold trace (gold spans for C-MTP, gold tokens for CoT) and let the model generate the remaining spans and the answer, sweeping $x$ from $0$ (original evaluation) to $1$ (only the answer is generated). Accuracy improves smoothly with $x$, and at $x=1$ all C-MTP variants nearly recover the discrete-CoT ceiling ($\sim$95/90/88\% at $2\times$/$3\times$/$4\times$ compression on NL, near-complete recovery on structured). To perform the sweep on structured traces, we use CoT generations filtered for correct answers as no structured ground truth exists for eval tasks.
    }}
    \label{fig:error_and_goldfill}
\end{figure}

\begin{table*}[t]
    \caption{\textbf{Training on Semi-Natural CoT: GSM8k-Aug-NL}. Accuracy is averaged over 5 runs (run-to-run deviation $<$1\%). \#L denotes the mean chain-of-thought length (in steps) with standard deviation, computed over the full evaluation set.}
    \centering
    \setlength{\tabcolsep}{2pt}
    \small
    \begin{tabular}{ll cc cc cc cc}
        \toprule
        Training Method & & \multicolumn{2}{c}{GSM8k-Test} & \multicolumn{2}{c}{GSM-Hard} & \multicolumn{2}{c}{SVAMP} & \multicolumn{2}{c}{MultiArith} \\
        \cmidrule(lr){3-4} \cmidrule(lr){5-6} \cmidrule(lr){7-8} \cmidrule(lr){9-10}
        & & Acc. ($\uparrow$) & \# L ($\downarrow$) & Acc. ($\uparrow$) & \# L ($\downarrow$) & Acc. ($\uparrow$) & \# L ($\downarrow$) & Acc. ($\uparrow$) & \# L ($\downarrow$) \\
        \midrule
        \rowcolor{gray!15} Standard & CoT & 53.68 & $70.04_{\pm 30.02}$ & 13.04 & $78.38_{\pm 33.01}$ & 60.88 & $36.77_{\pm 18.52}$ & 96.33 & $41.64_{\pm 10.82}$ \\
        \midrule
        Indirect & CODI & \textbf{47.82} & \textbf{6.00} & \textbf{10.99} & \textbf{6.00} & \textbf{56.82} & \textbf{6.00} & \textbf{95.56} & \textbf{6.00} \\
        Supervision & Sim-CoT & 44.35 & 6.00 & 10.11 & 6.00 & 55.16 & 6.00 & 97.89 & 6.00 \\
        \midrule
        Direct & CoLaR-2 & 23.46 & $35.63_{\pm 14.46}$ & 5.43 & $38.99_{\pm 15.02}$ & 37.70 & $19.15_{\pm 9.48}$ & 72.22 & $20.69_{\pm 7.59}$ \\
        Supervision & CoLaR-3 & 19.88 & $23.31_{\pm 10.01}$ & 4.61 & $25.55_{\pm 10.65}$ & 33.86 & $12.14_{\pm 5.56}$ & 67.67 & $13.86_{\pm 3.82}$ \\
        & CoLaR-4 & 18.06 & $18.15_{\pm 8.28}$ & 3.75 & $20.36_{\pm 9.33}$ & 33.16 & $9.43_{\pm 4.51}$ & 63.89 & $10.96_{\pm 4.68}$ \\
        \cmidrule(lr){2-10}
        & C-MTP-2 & 43.49 & $34.00_{\pm 15.04}$ & 10.13 & $37.32_{\pm 15.87}$ & 56.60 & $16.53_{\pm 8.83}$ & 91.22 & $18.74_{\pm 5.04}$ \\
        & C-MTP-3 & 34.04 & $22.74_{\pm 10.18}$ & 7.73 & $24.64_{\pm 10.78}$ & 48.24 & $10.57_{\pm 5.57}$ & 80.33 & $12.42_{\pm 3.48}$ \\
        & C-MTP-4 & 25.60 & $17.13_{\pm 7.79}$ & 6.20 & $19.00_{\pm 9.25}$ & 38.78 & $7.98_{\pm 4.16}$ & 67.11 & $9.26_{\pm 2.89}$ \\
        \bottomrule
    \end{tabular}
    \label{tab:trainnl}
\end{table*}

\begin{table*}[t]
    \caption{\textbf{Training on Realistic CoT: GSM-realistic}. Accuracy is averaged over 5 runs (run-to-run deviation $<$1\%). \#L denotes the mean chain-of-thought length (in steps) with standard deviation, computed over the full evaluation set.}
    \centering
    \setlength{\tabcolsep}{2pt}
    \small
    \begin{tabular}{ll cc cc cc cc}
        \toprule
        Training Method & & \multicolumn{2}{c}{GSM8k-Test} & \multicolumn{2}{c}{GSM-Hard} & \multicolumn{2}{c}{SVAMP} & \multicolumn{2}{c}{MultiArith} \\
        \cmidrule(lr){3-4} \cmidrule(lr){5-6} \cmidrule(lr){7-8} \cmidrule(lr){9-10}
        & & Acc. ($\uparrow$) & \# L ($\downarrow$) & Acc. ($\uparrow$) & \# L ($\downarrow$) & Acc. ($\uparrow$) & \# L ($\downarrow$) & Acc. ($\uparrow$) & \# L ($\downarrow$) \\
        \midrule
        \rowcolor{gray!15} Standard & CoT & 44.02 & $173.56_{\pm 65.36}$ & 11.04 & $191.00_{\pm 71.69}$ & 56.74 & $124.38_{\pm 40.38}$ & 83.44 & $116.33_{\pm 23.95}$ \\
        \midrule
        Indirect & CODI-6 & 25.46 & 6.00 & 6.32 & 6.00 & 40.84 & 6.00 & 57.11 & 6.00 \\
        Supervision & CODI-10 & 28.42 & 10.00 & 6.44 & 10.00 & 42.70 & 10.00 & 58.22 & 10.00 \\
        & CODI-20 & 27.54 & 20.00 & 6.63 & 20.00 & 41.62 & 20.00 & 62.33 & 20.00 \\
        \midrule
        Direct & C-MTP-2 & 35.21 & $86.09_{\pm 39.07}$ & 7.79 & $92.62_{\pm 42.57}$ & 46.86 & $61.23_{\pm 33.29}$ & 85.89 & $56.94_{\pm 18.92}$ \\
        Supervision & C-MTP-3 & 29.43 & $56.71_{\pm 21.47}$ & 6.31 & $60.08_{\pm 24.37}$ & 41.88 & $40.41_{\pm 19.78}$ & 78.00 & $39.40_{\pm 23.05}$ \\
        & C-MTP-4 & 23.38 & $42.25_{\pm 19.56}$ & 5.38 & $45.81_{\pm 30.49}$ & 36.16 & $30.21_{\pm 10.02}$ & 71.22 & $28.97_{\pm 6.89}$ \\
        \bottomrule
    \end{tabular}
    \label{tab:traingsmreal}
\end{table*}

\subsection{Training on Semi-Natural CoT: GSM8k-Aug-NL}
\label{sec:trainnl}

\cref{tab:trainnl} compares C-MTP and baselines on GSM8k-Aug-NL, which contains verbose natural language reasoning traces averaging $\sim$70 steps — roughly $2.5\times$ longer than GSM8k-Aug. \textbf{C-MTP outperforms CoLaR} across all compression ratios, consistent with \cref{sec:trainaug}. More strikingly, the gap between indirect and direct supervision widens considerably relative to the structured setting: CODI achieves 47.82\% on GSM8k-Test with just 6 latent steps, while CoLaR-2 collapses to 23.46\%, and even C-MTP-2, despite using $\sim$34 steps, reaches only 43.49\%.

Two factors, \rev{\change{13}both following from the regime definitions (\cref{app:definitionexplain})}, explain why verbose data disproportionately hurts parallel methods. First, indirect supervision is inherently suited to compression: a fixed latent budget forces the model to discard filler tokens and irrelevant content, whereas parallel methods must predict the full verbose trace without any such pressure. Second, teacher forcing causes a train-test mismatch: at inference, the model conditions on its own prior predictions rather than ground-truth CoT traces, with errors cascading particularly severely through math expressions. As we see in the realistic setting, this effect is especially pronounced on math expressions where a single erroneous token can affect the accuracy of the entire CoT trace.

\subsection{Analysis of C-MTP Generations}

\paragraph{Mathematical validity in structured outputs.} Training on GSM8k-Aug teaches models to reason using only expressions relevant to computing the final answer. We compute the mathematical validity of each generated expression for each sample and plot the average fraction of valid expressions at each position in the generated CoT trace for both C-MTP and CoT in \cref{fig:experroraccum}.  We see that C-MTP starts with a high fraction at the beginning of the trace and falls off towards the end, especially for $s=4$ where the model is forced to compress more. We further analyze generations upon training on GSM8k-Aug-NL, where the traces contain both math expressions and natural language. We find that the model gets things wrong primarily in math expressions (see examples in \cref{app:mtperrorsexp}) especially when a span boundary occurs in between. Upon an incorrect generation the model is forced to condition on its own erroneous output (error accumulation), which leads to an incorrect answer. We detail a few selected examples in \cref{app:mtperrors}.

\rev{\change{4}
\paragraph{Gold trace prefill.} To separate errors in generated spans from the loss due to token aggregation, we prefill the trained C-MTP model with a fraction of oracle ground truth spans and then let it generate remaining spans and the final answer. We sweep $x$ from $0$ (original evaluation) to $1$ (the entire gold trace is fed and only the answer is generated). For C-MTP, $x$ is the fraction of gold \emph{spans} whereas for CoT, $x$ is the fraction of gold tokens. For structured tasks, we rely on correct CoT generations for traces as ground truth traces do not exist for test data. $x=1$ measures the attainable ceiling where all thinking steps are correct and shows the gap due to aggregation of tokens into single span. \cref{fig:gsmstructgsm_goldfill,fig:gsmnl_goldfill} show that all C-MTP variants nearly recover the discrete CoT ceiling at $x=1$. C-MTP achieves ~95\%,~90\% and ~88\% score with 2$\times$,3$\times$ and 4$\times$ compression respectively on NL tasks. On structured tasks, C-MTP variants recover nearly all of the performance. Mean pooling therefore accounts for only a small error residual. There is a smooth improvement in performance as we increase the fraction of gold trace. Prediction errors and their propagation both contribute to the gap between curves at $x<1$.
}

\paragraph{Experiments with pure-NL task.} We further experiment with CommonsenseQA \citep{talmor-etal-2019-commonsenseqa}, a commonsense multiple choice question answering dataset paired with natural language CoTs generated by GPT-4o-mini \citep{DBLP:conf/emnlp/ShenYZHDH25} in \cref{app:commonsenseexps}. Interestingly, we do not see a severe degradation of C-MTP with increasing span lengths, validating our observation that errors cascade mainly through precise tokens such as math expressions.

\begin{table}[!t]
    \caption{Results on MATH across two backbones. SD=0.0 indicates that the model performs best using greedy decoding.}
    \label{tab:trainmath}
    \centering
    \footnotesize
    \setlength{\tabcolsep}{2pt}
    \begin{minipage}[t]{0.54\textwidth}
        \centering
        \textit{LLaMA-3.2-1B-Instruct}\par\vspace{2pt}
        \begin{tabular}{ll cc}
            \toprule
            Training & & Acc. ($\uparrow$) & \# L ($\downarrow$) \\
            \midrule
            \rowcolor{gray!15} & Original & $24.00_{\pm 0.00}$ & $826.60_{\pm 1464.90}$ \\
            \rowcolor{gray!15} Standard & CoT-SFT & $23.80_{\pm 0.00}$ & $350.10_{\pm 271.15}$ \\
            \midrule
            \multirow{4}{*}{Direct} & C-MTP-2 & $7.40_{\pm 0.00}$ & $163.62_{\pm 106.74}$ \\
            & \shortstack[l]{CoLaR-2 \\ (reproduced)} & $4.36_{\pm 0.88}$ & $129.71_{\pm 65.49}$ \\
            & (reported) & $5.28_{\pm 0.16}$ & $83.10_{\pm 0.16}$ \\
            & (reported SFT+RL) & $7.08_{\pm 0.00}$ & $16.10_{\pm 0.14}$ \\
            \midrule
            \multirow{3}{*}{Indirect} & CODI-6  & $8.52_{\pm 0.20}$ & 6.00 \\
             & CODI-10 & $9.92_{\pm 0.20}$ & 10.00 \\
             & CODI-20 & $7.92_{\pm 0.16}$  & 20.00 \\
            \bottomrule
        \end{tabular}
    \end{minipage}\hfill
    \begin{minipage}[t]{0.44\textwidth}
        \centering
        \textit{Qwen2.5-1.5B-Instruct}\par\vspace{2pt}
        \begin{tabular}{ll cc}
            \toprule
            Training & & Acc. ($\uparrow$) & \# L ($\downarrow$) \\
            \midrule
            \rowcolor{gray!15} & Original & $53.52_{\pm 0.70}$ & $644.90_{\pm 460.30}$ \\
            \rowcolor{gray!15} Standard & CoT-SFT & $52.80_{\pm 0.00}$ & $547.37_{\pm 281.63}$ \\
            \midrule
            \multirow{2}{*}{Direct} & C-MTP-2 & $14.20_{\pm 0.00}$ & $272.11_{\pm 124.83}$ \\
            & \shortstack[l]{CoLaR-2 \\ (reproduced)} & $5.28_{\pm 0.52}$ & $202.11_{\pm 3.26}$ \\
            \midrule
            \multirow{3}{*}{Indirect} & CODI-6  & $12.44_{\pm 0.15}$ & 6.00 \\
             & CODI-10 & $14.52_{\pm 0.20}$ & 10.00 \\
             & CODI-20 & $12.96_{\pm 0.50}$ & 20.00 \\
            \bottomrule
        \end{tabular}
    \end{minipage}
\end{table}

\subsection{Training on Realistic CoT: \rev{\change{1}GSM-realistic and} MATH}
\label{sec:trainmath}

\rev{\change{1}To evaluate how direct and indirect supervision scale to longer and more complex CoT traces, we train and evaluate both regimes on (a) GSM-realistic, a subset of GSM8k-Aug problems with longer, more verbose answers and (b) MATH, a challenging dataset of high school level math problems}. For both of these sets, we generate answers to all problems in the train set, filter out wrong answers and use the correct generations for training. \rev{\change{8}We provide more details on our data generation procedure in \cref{app:realisticdetails}}.

\subsubsection{GSM-realistic}

\rev{\change{1}We finetune CoT-SFT, C-MTP-2/3/4 and CODI-6/10/20 to compare discrete CoT, direct and indirect supervision methods. Table \ref{tab:traingsmreal} shows two main findings: (1) The gap between CoT-SFT and the best performing direct and indirect supervision methods becomes more pronounced ($\sim$18\% drop). C-MTP-2 achieves 35.21\% accuracy with $\sim$86 steps, while CODI-10 achieves 28.42\% with just 10 latent steps, compared to 44.02\% for CoT-SFT with $\sim$174 steps. (2) C-MTP-2 outperforms CODI, reversing the trend from the semi-natural setting, suggesting that the compression ability of indirect supervision is bounded and does not always scale to longer traces. }

\subsubsection{MATH}
Upon moving to more complex and verbose MATH CoT traces, we see that the gaps between CoT-SFT and both direct and indirect supervision methods widen even further. In \cref{tab:trainmath}, we see that both methods drastically fail, achieving only $\sim$12-15\% accuracy compared to $\sim$53\% for CoT-SFT for Qwen2.5-1.5B-Instruct. Experiments with LLaMA show similar trends with all methods achieving <10\% compared to 23.8\% for CoT-SFT. \rev{\change{6}CoLaR \citep{tan2026think} reports 7.08\% accuracy on MATH with SFT+RL, showing that RL is highly effective for reducing output lengths ($\sim 5\times$ decrease) but does not close the accuracy gap}.

Although CODI's reasoning traces are not interpretable, analysis of traces generated by C-MTP highlights similar issues of error accumulation seen in semi-natural tasks \rev{\change{4} (\cref{fig:error_and_goldfill})}. The model generates the right natural language tokens but makes arithmetic errors, drops signs or numbers or produces malformed expressions, which leads to an irrecoverable error in the trace and an incorrect final answer, suggesting that future ability to improve such methods may require a combination of the two. Direct supervision may be essential for its efficiency and alignment with the token-based training objective and indirect supervision may be essential for its ability to compress long traces into fixed latent budgets and avoid train-test mismatch.

\section{Discussion}
\rev{\change{3,13}
Overall our experiments in \cref{sec:trainaug,sec:trainnl,sec:trainmath} show a clear trend in the relative effectiveness of the two regimes as the length and complexity of the reasoning traces increase. We note that our discussion below applies to the methods we evaluate and SFT trained models in 1-1.5B size range (\cref{sec:limitations}).

On short and compact structured traces, direct supervision performs comparably to indirect supervision, with the best method trailing CoT-SFT by only $\sim$9\%. The two regimes differ in inference efficiency, which we estimate by the number of reasoning steps since C-MTP adds negligible measured per-step latency over a standard forward pass (\cref{app:mtpoverhead}). On problems that require short traces (e.g SVAMP), variable-length C-MTP outperforms fixed-length CODI while using fewer steps (62.10\% with 5.14 steps vs.\ 60.26\% with 6 steps), whereas on tasks that require longer traces (e.g., GSM8k-Test), C-MTP is less efficient than CODI (53.30\% with 12.07 steps vs.\ 53.03\% with 6 steps; \cref{tab:trainaugllama}). This difference arises because indirect methods fix the number of latent steps prior to training, whereas direct methods fix a compression factor which ties their generation length to problem type (\cref{tab:supcomparison}). Varying the compression factor $s$ traces out the step-accuracy curve, with C-MTP-1 recovering standard CoT by construction. Approaches such as \citet{ning2025learningstopadaptivelatent} seek to make CODI adaptive to problem difficulty, a property that direct supervision provides naturally. Considering the additional benefits such as sample efficiency, faster training and better generalization to held-out problem structures, direct supervision remains the preferable choice in this setting.

The semi-natural trace setting holds the problem difficulty fixed and increases the length and verbosity of the CoT traces exposing the weaknesses of direct supervision methods. Our analysis of C-MTP in \cref{sec:trainnl} and \cref{fig:error_and_goldfill} shows that errors in generated spans, rather than token aggregation, account for most of its degraded performance. Prediction errors and their propagation account for the performance gap from CoT at $x<1$. CODI, in contrast, compresses the filler content effectively and trails CoT-SFT by only $\sim$9\%, nearly the same gap as on structured traces. The GSM-realistic task, trained on the GSM8k-Aug problems the model answers correctly, increases trace length further, stress-testing the compression ability of indirect supervision. Here CODI degrades noticeably, and while C-MTP now outperforms CODI, it still trails CoT-SFT by $\sim$18\%, indicating that both regimes begin to regress. Finally, the MATH setting increases both the problem difficulty and trace length resulting in both regimes nearly collapsing, with the best latent configuration trailing CoT-SFT by $\sim$60\% on LLaMA-3.2-1B-Instruct. The two regimes attain comparable accuracy (7.40\% for C-MTP-2 vs.\ 9.92\% for CODI-10), even though C-MTP-2 uses roughly $16\times$ more steps (\cref{tab:trainmath}). For CODI, the very property that makes it more efficient on structured and semi-natural traces, namely compression into a fixed latent budget, starts becoming a liability on longer and more complex traces that are difficult to compress.

Taken together, these results show that structured traces used predominantly in prior works \citep{wei2026simcot,DBLP:journals/corr/abs-2606-31779} do not capture all characteristics of latent reasoning methods. Advantages displayed in one setting do not always reliably carry over as traces grow longer and problems become harder. Since neither regime alone closes the gap to CoT-SFT on realistic traces, our findings underscore the need for approaches that combine the strengths of both (\cref{sec:limitations}).

}

\section{Limitations and Future Work}
\label{sec:limitations}
Our regime-level classification and conclusions are drawn from the methods we primarily consider (C-MTP and CoLaR for direct supervision; CODI, Coconut, and Sim-CoT for indirect supervision) and SFT-trained models in 1-1.5B size range. We refer to this scope as our \emph{evaluated setting} in our work. Our work focuses on mathematical reasoning tasks, which may limit the generality of our findings. We focus on math because prior work has shown that latent reasoning methods struggle to match CoT in this domain, making it a natural testbed. Extending the evaluation of C-MTP to tasks such as code and multi-hop question answering with larger scale models is a natural next step.  More fundamentally, our realistic-trace results suggest that neither direct nor indirect supervision alone suffices: a promising direction is hybrid approaches that adaptively switch between token-based and continuous reasoning \citep{shi2026swireasoningswitchthinkinglatentexplicit, xu2026thinkrouterefficientreasoningrouting, zhu2026lightthinkerreasoningcompressionmemory}, which may be essential for scaling ContinuousCoT to realistic settings. \rev{\change{7}We note that our experiments are restricted to the supervised finetuning setting. Reinforcement learning is a complementary improvement approach that can be applied to C-MTP as well. Exploring both regimes with RL is a natural direction for future work. \change{17} We also note that variational and/or diffusion based objectives are a promising direction for latent reasoning that do not fit cleanly into the two regimes.}

\begin{figure*}[t]
    \centering
    \includegraphics[width=\textwidth]{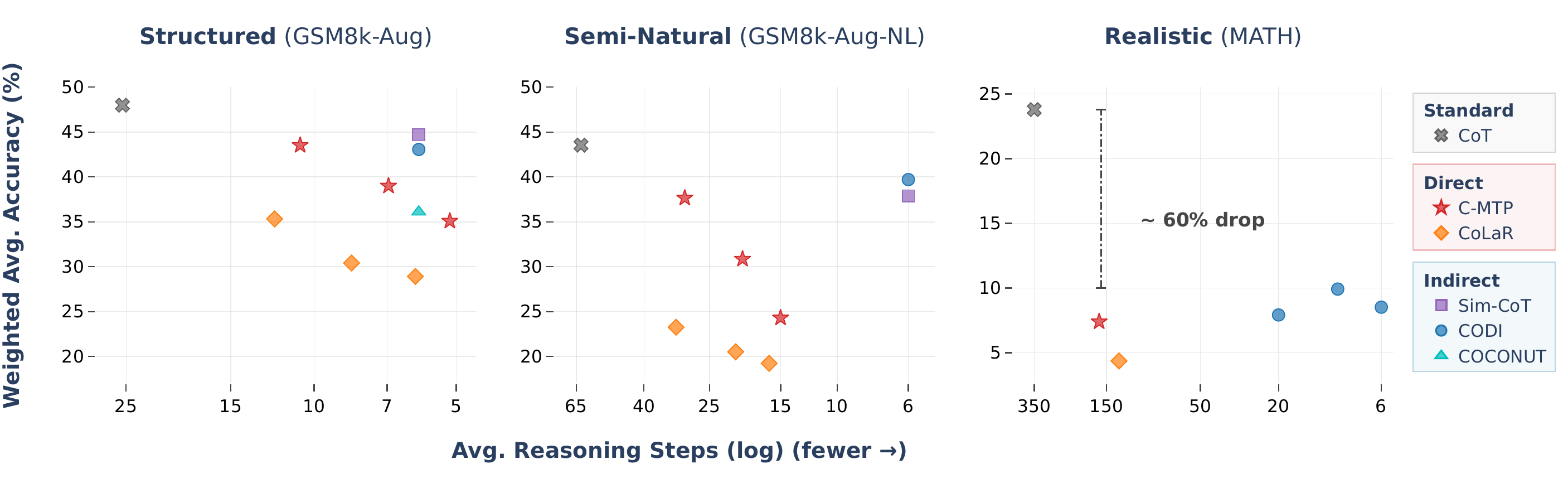}
    \label{fig:allresults}
    \vspace{-1em}
    \caption{\textbf{LLaMA-3.2-1B-Instruct Results at a glance.} A comparative analysis of weighted average accuracy (\%) versus the average number of reasoning steps (log scale) for Standard CoT and two regimes (direct and indirect) of ContinuousCoT across three levels of trace difficulty. On simpler tasks (structured and semi-natural), both regimes perform competitively with Standard CoT. However, on complex, realistic tasks, even the best configuration reaches only $\sim$40\% of CoT accuracy (\cref{tab:summarydrops}). This demonstrates that simplified benchmarks overstate the practical effectiveness of continuous CoT models in realistic settings.}
\end{figure*}

\section{Conclusion}

We propose C-MTP, a direct supervision method using multi-token prediction to simulate ContinuousCoT, which substantially outperforms comparable approaches. Through systematic analysis of direct and indirect supervision across Structured, Semi-Natural, and Realistic CoT traces, we find within our evaluated setting (\cref{sec:limitations}): direct supervision (C-MTP) is preferable on structured tasks for its efficiency; indirect supervision dominates on semi-natural tasks, better compressing long traces into fixed latent budgets where direct methods fail and face train-test mismatch; and on realistic traces, both regimes fall far short of CoT-SFT (\cref{fig:mainresults}). Our results suggest that structured and semi-natural training significantly overstates the effectiveness of Continuous CoT methods.

\section*{Acknowledgements}
This work was supported in part through the NYU IT High Performance Computing resources, services, and staff expertise. The work is partially funded by NSF CAREER award 2443271 and NSF award RI-2521091.

\rev{\change{15}
\section*{Broader Impact Statement}
Research on replacing explicit reasoning traces with latent representations can potentially reduce interpretability and auditability of a model's intermediate reasoning making it harder to inspect, verify or monitor the model. While several activation based approaches aim to capture intent of LLMs before output, direct supervision sits at the more inspectable end of the spectrum of latent reasoning approaches. C-MTP, to be concrete, constrains each latent to an aggregation of token embeddings, so its intermediate steps can be read back as the tokens predicted at each step, unlike the unconstrained continuous latents of indirect supervision. We encourage practitioners deploying continuous CoT models in settings where reasoning must be audited to weigh this tradeoff between inference efficiency and transparency.
}

\vfill

\newpage
\clearpage

\bibliography{main}
\bibliographystyle{tmlr}

\appendix
\crefalias{section}{appendix}
\crefalias{subsection}{subappendix}
\onecolumn

\vspace{10em}

\section*{\Large Table of Contents} \vspace{1em}
{\large
\begin{itemize}
    \item[\textbf{A.}] MTP Architecture
    {\normalsize\begin{itemize}
        \item[\textbf{A.1}] \rev{\change{11}Inference Overhead of Multi-Token Prediction}
    \end{itemize}}
    \item[\textbf{B.}] \rev{\change{12,13}Properties of Direct and Indirect Supervision}
    \item[\textbf{C.}] Example of CoT Traces for Three Difficulties
    \item[\textbf{D.}] Timing Analysis for Fixed Wallclock Time Experiments
    \item[\textbf{E.}] Analysis of Problem Structures in GSM8k-Aug
    \item[\textbf{F.}] Examples of Error Accumulation during MTP Inference
    \item[\textbf{G.}] C-MTP Ablation Studies
    {\normalsize\begin{itemize}
        \item[\textbf{G.1}] \mttrain{} vs \wsinit{}
        \item[\textbf{G.2}] Ablation on C-MTP: Training without $L_{KD}$
    \end{itemize}}
    \item[\textbf{H.}] Extended Experimental Results
    {\normalsize\begin{itemize}
        \item[\textbf{H.1}] Setup Details
        \item[\textbf{H.2}] Training on Structured CoT: GSM8k-Aug
        \item[\textbf{H.3}] Sample Efficiency
        \item[\textbf{H.4}] Fixed Wallclock Time Comparison
        \item[\textbf{H.5}] Generalization to Problem Structures
        \item[\textbf{H.6}] Gold Trace Analysis
        \item[\textbf{H.7}] Ablation Studies
        \item[\textbf{H.8}] Full Example of MTP Error Accumulation
        \item[\textbf{H.9}] Experiments on CommonsenseQA
    \end{itemize}}
    \item[\textbf{I.}] \rev{\change{8}Details on Realistic CoT Datasets}
    \item[\textbf{J.}] CoLaR - CODI Implementation Discrepancy
\end{itemize}
}
\newpage

\section{MTP Architecture}
\label{app:mtp-arch}

Standard LLMs generate a single probability distribution over $\vocabV$. To generate multiple distributions from the same backbone, we train a small MLP to project the final layer activation into activations corresponding to multiple tokens. We then use the same output embedding matrix to generate logits for all tokens.

At each thinking step $i$, the hidden state is obtained by passing the
prompt and all predicted latents through the model:
\begin{align}
    h_i = H_\theta(E(Q),\, z_1, \ldots, z_{i-1}) \in \realR^d
\end{align}
A 2-layer MLP: $g \in \realR^{d \times sd}$ maps this to $s$ token
representations simultaneously:
\begin{align}
    O_i = \mathrm{reshape}(g(h_i)) \in \realR^{s \times d}
\end{align}
Row $O_i[j]$ is the predicted representation for the $j$-th token of the
next span $C_{i+1}$. Logits are obtained by projecting through the output embedding matrix $E \in \realR^{|\vocabV| \times d}$:
\begin{align}
    P_\theta(C_{i+1}[j] \mid E(Q), z_{1:i})
    = \mathrm{softmax}\!\left( O_i[j] \cdot E' \;\right)
\end{align}
A single forward pass thus produces all $s$ token distributions in parallel, with no sequential decoding within a span. We then sample one token from each distribution to form the next span $C_{i+1}$, and aggregate their embeddings to obtain the next latent $z_{i+1}$ via \cref{eqn:embagg}. \rev{\change{5}With this setup, the ordering of tokens is preserved on the output side. C-MTP averages input embeddings which is permutation-invariant and loses the positional information. During our preliminary experiments, a RoPE-style rotation before averaging did not improve validation or test accuracy, indicating that preserving positional information while compression remains an open problem.}

\rev{\change{11}
\subsection{Inference Overhead of Multi-Token Prediction}
\label{app:mtpoverhead}

Since C-MTP shares the output embedding matrix across all tokens, the only additional cost over standard NTP is the MLP projection $g$ of the final layer activation into $s$ token representations. This can be batched, resulting in a very small overhead. To verify this, we measure latency for both the \emph{full forward step} (the entire transformer pass that produces the next latent) and the \emph{prediction head only} (the projection and batched embedding matrix multiplication, the component that scales with the number of predicted tokens). \cref{tab:mtpoverhead} (Appendix) reports results with batch size 32 and sequence length 180, averaged over repeated runs.

\begin{table}[h]
    \centering
    \caption{\textbf{Inference latency of a single decoding step for NTP and C-MTP}. \textit{Full step}: entire transformer forward pass producing the next latent. \textit{Head only}: projection and batched embedding matrix multiplication. Batch size 32, sequence length 180, mean $\pm$ std over repeated runs.}
    \begin{tabular}{llccc}
        \toprule
        Model & Method & Full step (ms) & Head only (ms) & Head overhead vs.\ NTP (ms) \\
        \midrule
        \multirow{3}{*}{\shortstack[l]{LLaMA-3.2-1B \\ ($d$ = 2048, $|\vocabV|$ = 128256)}} & NTP & $34.02 \pm 0.6$ & $0.228 \pm 0.0$ & -- \\
        & C-MTP-2 & $34.02 \pm 0.2$ & $0.247 \pm 0.0$ & $+0.019$ \\
        & C-MTP-3 & $34.10 \pm 0.2$ & $0.240 \pm 0.0$ & $+0.012$ \\
        \midrule
        \multirow{3}{*}{\shortstack[l]{Qwen2.5-1.5B \\ ($d$ = 1536, $|\vocabV|$ = 151936)}} & NTP & $44.35 \pm 0.8$ & $0.204 \pm 0.0$ & -- \\
        & C-MTP-2 & $44.38 \pm 0.8$ & $0.221 \pm 0.0$ & $+0.017$ \\
        & C-MTP-3 & $44.50 \pm 0.9$ & $0.228 \pm 0.0$ & $+0.024$ \\
        \bottomrule
    \end{tabular}
    \label{tab:mtpoverhead}
\end{table}

We see that predicting 2 or 3 tokens per step leaves the full step latency essentially unchanged (LLaMA: 34.02 $\rightarrow$ 34.10 ms; Qwen: 44.35 $\rightarrow$ 44.50 ms; $<0.4\%$). Since the two architectures differ only in how the final layer activation and embeddings are used, we estimate end-to-end inference cost relative to CODI by the number of reasoning steps: C-MTP is expected to be faster than CODI when it uses fewer steps than CODI's fixed latent budget, and slower otherwise.
}

\rev{\change{12,13}
\section{Properties of Direct and Indirect Supervision}
\label{app:definitionexplain}

The properties listed in \cref{tab:supcomparison} follow from a single defining difference between the regimes: under \textbf{indirect supervision}, the latents assume no particular structure and are supervised only through a downstream objective, whereas under \textbf{direct supervision}, the CoT trace \emph{defines} the latents.

\paragraph{Output (continuous vs. discrete).} Under direct supervision, each latent is tied to specific tokens of the trace, which constrains latents to the subspace spanned by (aggregates of) token embeddings, so the outputs are effectively discrete. Under indirect supervision, a latent can be any continuous representation the architecture can produce. For example, although CoLaR \citep{tan2026think} predicts a Gaussian latent, it is trained to approximate a distribution over the tokens within each compressed span, i.e., token-grounded targets with the Gaussian acting as a reparameterization.

\paragraph{Reasoning length (fixed vs. variable).} Under direct supervision, the number of latents is determined by the token length of the trace (e.g., $\lceil |y| / s \rceil$ for C-MTP-$s$), and is therefore variable. Under indirect supervision, the number of latent steps can be set independently of the trace length, and is fixed before training.

\paragraph{Training method and efficiency (recurrence vs. teacher forcing).} Under direct supervision, the target at every position is known beforehand from the trace, so all positions can be trained in parallel with teacher forcing in a single forward pass. Under indirect supervision, the input to step $t$ is the model's own latent from step $t-1$, which is not available in advance, so training step $t$ requires rolling the forward pass through steps $1, \ldots, t-1$, making training recurrent by construction. We provide actual latencies in \cref{sec:timing_analysis}.

Within our evaluated setting (\cref{sec:limitations}), multiple results we observe in \cref{sec:trainnl} follow from these properties. Indirect supervision ends up compressing verbose traces better because its latent budget is fixed independently of the trace, it is explicitly trained to pack the reasoning content into a small number of latents regardless of the verbosity of the trace. This advantage is not unbounded as we see in \cref{sec:trainmath}. Direct supervision is more prone to error accumulation due to teacher forcing as it is trained on ground truth spans but conditions on its own generated spans at inference, which CODI (indirect supervision) does not. This training--inference mismatch compounds small errors along the trace, and its effect grows with trace length. Consistent with this, accuracy recovers as more ground-truth spans are provided (\cref{fig:error_and_goldfill})

\change{14} In our evaluated setting, direct supervision methods predict either an $R^d$ vector composed of token embeddings (CoLaR) or explicit tokens that are then aggregated into $R^d$. The subsequent thinking steps condition on these \emph{dense} vectors providing efficiency benefits as expected from a ContinuousCoT method. Since the latent structure primarily comprises of \emph{tokens}, learning to reason in this space is much simpler and closer to the original model's distribution resulting in better sample efficiency and generalization in structured tasks.
}

\section{Example of CoT Traces for Three Difficulties}
\label{app:exampletraces}

Bella bought 18 packs of chocolates each containing 12 chocolates. If Bella gives away 6 packs, how many dozens of chocolates will she have left?

\begin{table}[htbp]
    \centering
    \small
    \renewcommand{\arraystretch}{1.5}
    \begin{tabular}{|p{0.2\textwidth}|p{0.3\textwidth}|p{0.4\textwidth}|}
        \hline
        \centering\textbf{Structured} & \centering\textbf{Semi-Natural} & \centering\arraybackslash\textbf{Realistic} \\
        \hline

        $\ll$18*12=216$\gg$ \newline
        $\ll$216-6*12=144$\gg$ \newline
        $\ll$144/12=12$\gg$

        &
        Bella initially had 18 packs * 12 chocolates/pack = 216 chocolates. \vspace{\baselineskip} \newline
        After giving away chocolates, Bella has 216 chocolates - 6 packs * 12 chocolates/pack = 144 chocolates left. \vspace{\baselineskip} \newline
        Thus, Bella is left with 144 chocolates / 12 = 12 dozens of chocolates.

        &

        \footnotesize
        To determine how many dozens of chocolates Bella has left after giving some away, we can follow these steps: \vspace{2mm} \newline
        \textbf{1. Calculate the total number of chocolates Bella initially had:} \newline
        - Each pack contains 12 chocolates. \newline
        - Bella bought 18 packs. \newline
        - Therefore, the total number of chocolates is: \newline
        $18 \text{ packs} \times 12 \text{ chocolates per pack} = 216 \text{ chocolates}$ \vspace{2mm} \newline
        \textbf{2. Determine the number of chocolates in the packs Bella gave away:} \newline
        - Bella gave away 6 packs. \newline
        - Each pack contains 12 chocolates. \newline
        - Therefore, the number of chocolates given away is: \newline
        $6 \text{ packs} \times 12 \text{ chocolates per pack} = 72 \text{ chocolates}$ \vspace{2mm} \newline
        \textbf{3. Calculate the number of chocolates remaining:} \newline
        - Subtract the number of chocolates given away from the initial total: \newline
        $216 \text{ chocolates} - 72 \text{ chocolates} = 144 \text{ chocolates}$ \vspace{2mm} \newline
        \textbf{4. Convert the remaining chocolates into dozens:} \newline
        - There are 12 chocolates in a dozen. \newline
        - To find out how many dozens of chocolates remain, divide the number of chocolates by 12: \newline
        \textbackslash{}frac\{144 \textbackslash{}text\{ chocolates\}\}\{12 \textbackslash{}text\{ chocolates per dozen\}\} = 12 \textbackslash{}text\{ dozens\} \vspace{2mm} \newline
        Thus, the number of dozens of chocolates Bella has left is \textbackslash{}(\textbackslash{}boxed\{12\}\textbackslash{}) \\

        \hline
    \end{tabular}
    \normalsize
    \caption{CoT Trace Examples}
    \label{tab:sampletraces}
\end{table}

Realistic Trace generated from Qwen2.5-1.5B-Instruct.

\newpage
\section{Timing Analysis for Fixed Wallclock Time Experiments}
\label{sec:timing_analysis}

To ensure a fair comparison for our fixed wallclock time experiments, we measure the latency of a single training step (one batch forward and backward) for both architectures. We benchmark our direct supervision method (MTP) using a span length of 4 against indirect supervision (CODI) using 6 latent steps. These configurations were selected as they result in similar compression ratios on the GSM8k-Aug dataset. All timing evaluations are conducted on an H200 GPU, with each sequence partitioned into a 60-30-10 ratio token split for the question, CoT, and answer, respectively. Model configurations are detailed in \cref{tab:timing_analysis} (Appendix).

\rev{\change{9}
\begin{table}[h]
    \centering
    \caption{Latency (ms) of a single training step (forward + backward) for C-MTP-4 and CODI-6 on LLaMA-3.2-1B-Instruct, mean over 20 trials after 5 warmup steps. Speedup is CODI latency / C-MTP latency ($>1$ means C-MTP is faster). C-MTP uses full-logit materialization for all MTP heads.}
    \small
    \setlength{\tabcolsep}{4pt}
    \begin{tabular}{r ccc ccc ccc}
        \toprule
        & \multicolumn{3}{c}{Batch size 16} & \multicolumn{3}{c}{Batch size 32} & \multicolumn{3}{c}{Batch size 64} \\
        \cmidrule(lr){2-4} \cmidrule(lr){5-7} \cmidrule(lr){8-10}
        Seq.\ Len. & C-MTP & CODI & Speedup & C-MTP & CODI & Speedup & C-MTP & CODI & Speedup \\
        \midrule
        32 & 161.30 & 639.81 & $3.97\times$ & 175.18 & 610.00 & $3.48\times$ & 228.95 & 660.58 & $2.89\times$ \\
        64 & 179.92 & 633.37 & $3.52\times$ & 224.79 & 615.10 & $2.74\times$ & 386.14 & 705.59 & $1.83\times$ \\
        96 & 199.02 & 637.82 & $3.20\times$ & 290.19 & 623.80 & $2.15\times$ & 544.23 & 791.40 & $1.45\times$ \\
        128 & 225.62 & 635.21 & $2.82\times$ & 381.28 & 661.60 & $1.74\times$ & 722.30 & 880.25 & $1.22\times$ \\
        \bottomrule
    \end{tabular}
    \label{tab:timing_analysis}
\end{table}

\begin{table}[h]
    \centering
    \caption{Using a slightly more memory-efficient implementation of MTP \citep{DBLP:conf/icml/GloeckleIRLS24} improves speedups in certain settings.}
    \small
    \setlength{\tabcolsep}{4pt}
    \begin{tabular}{c cccc c cccc}
        \toprule
        & \multicolumn{4}{c}{Full-logit MTP} & & \multicolumn{4}{c}{Per-head MTP} \\
        \cmidrule{2-5} \cmidrule{7-10}
        Batch \textbackslash\ Seq.\ Len. & 32 & 64 & 96 & 128 & & 32 & 64 & 96 & 128 \\
        \midrule
        16 & 3.97 & 3.52 & 3.20 & 2.82 & & 3.91 & 3.27 & 3.17 & 2.88 \\
        32 & 3.48 & 2.74 & 2.15 & 1.74 & & 3.46 & 2.83 & 2.25 & 1.85 \\
        64 & 2.89 & 1.83 & 1.45 & 1.22 & & 3.05 & 1.97 & 1.56 & 1.31 \\
        \bottomrule
    \end{tabular}
    \label{tab:timing_mtpimpl}
\end{table}

As we see in \cref{tab:timing_analysis} (Appendix), speedup of C-MTP over CODI depends strongly on batch size and sequence length. C-MTP's advantage comes from computing the loss for all positions in a single parallel pass, whereas CODI requires one sequential forward pass per latent step. While the total FLOPs are similar, the fully parallel pass achieves much better GPU utilization. At batch size 16, CODI's latency stays nearly flat across sequence lengths (633--640 ms), indicating that it is dominated by its sequential latent passes rather than by compute, while C-MTP's latency grows steadily with sequence length (161--226 ms). The speedup is therefore largest at small batch sizes and short sequences ($3.97\times$ at batch size 16, sequence length 32), which is the relevant regime for structured training traces (GSM8k-Aug traces have $21.46 \pm 10.46$ tokens). As batch size and sequence length grow, the MTP logit computation becomes compute-bound and the speedup shrinks. \cref{tab:timing_mtpimpl} (Appendix) shows that this bottleneck can be partially mitigated by the per-head implementation of \citet{DBLP:conf/icml/GloeckleIRLS24}, helping the most in demanding configurations (e.g., $1.22\times \rightarrow 1.31\times$ at batch size 64, sequence length 128).

Fully chunked logit-loss computation could improve C-MTP further whereas more efficient train-time KV-caching methods can improve CODI further. Neither method is optimized to its theoretical best while the measured training time is also sensitive to implementation details (compilation, sequence length, hardware, optimizer configuration, gradient accumulation). We therefore treat the $2\times$ speedup used in our fixed wallclock experiments (\cref{app:fixwallclock}) as a conservative estimate of the speedup, while the actual speedup on structured tasks can be much higher.
}

\section{Analysis of Problem Structures in GSM8k-Aug}
\label{app:gsmauganal}

As the dataset consists of expressions in the CoT traces, we parse the expressions into templates to analyze the diversity of problem structures. Some examples of templates are given below:

Count in Train: 24353  \textbf{Template: a/b=r0} \\
 Q: A school has 3 classrooms, each with the same number of students. If there are 36 students in total, how many students are there in each classroom? \\
 CoT: $\ll$36/3=12$\gg$

Count in Train: 22451  \textbf{Template: a*b=r0} \\
Q: Jacob is building a dog house. He uses 8 nails for each plank and needs 6 planks. How many nails will he need in total? \\
CoT: $\ll$8*6=48$\gg$

Count in Train: 14989  \textbf{Template: a*b=r0, r0*c=r1} \\
 Q: A basketball team consists of 12 players. Each player needs two basketball shoes. The cost of each shoe is \$75. How much will it cost for the basketball team to buy shoes? \\
 CoT: $\ll$12*2=24$\gg$ $\ll$24*75=1800$\gg$

\begin{figure}[t]
    \centering
    \caption{Cumulative coverage of top-n problem structures in GSM8k-Aug}
    \includegraphics[width=0.5\textwidth]{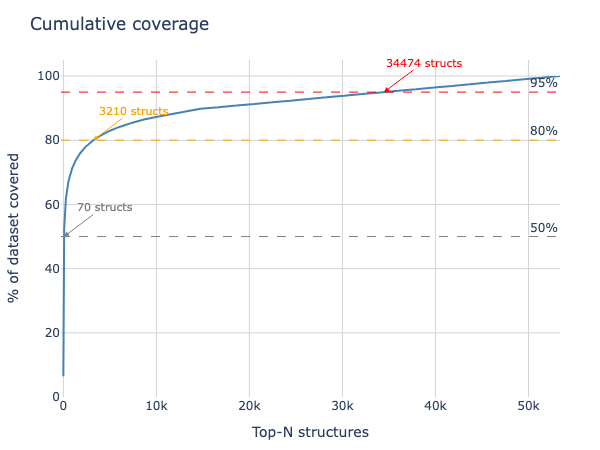}
    \label{fig:structurescoverage}
\end{figure}

 Cumulative coverage \cref{fig:structurescoverage} (Appendix) shows that 80\% of the dataset consists of 3k structures which is less than 1\% of the size of the dataset.

We find that the top 10, 50, 100, 500 structures cover 28.3\%, 46.5\%, 53.3\% and 66.5\% of the 380k samples of GSM8k-Aug which is a considerable portion. We hence argue that this dataset, does not reflect real-world diversity of typical SFT.

\section{Examples of Error Accumulation during MTP Inference}
\label{app:mtperrorsexp}

A collection of all wrong arithmetic expressions on Semi-Natural CoT

{\small

500000*.150=150 \;;\; \textbf{3*3=99} \;;\; 75-15=15 \;;\; 20-1=10 \;;\; 1010*1=10 \;;\; 16/3=3 \;;\; 20+2=60 \;;\; 4*4=160 \;;\; 10.2*10=24 \;;\; 24*45=108108 \;;\; 30/100=54 \;;\; 180+324=324 \;;\; 180-0.3=120 \;;\; 180+120=360 \;;\; 2*8080=160 \;;\; 6*55=330330 \;;\; 90/4=9 \;;\; 7*7=1 \;;\; 90/31=2 \;;\; 3*3=90 \;;\; 25/100=4 \;;\; 16-4=8 \;;\; 20*20.2=4 \;;\; 2020-4=16 \;;\; 8000*1.2=960 \;;\; \textbf{21/12=1} \;;\; 1.1*75=4 \;;\; 2+4=10 \;;\; 6/44=1 \;;\; 2/3=6 \;;\; 3/5=9 \;;\; .67+9=15 \;;\; \textbf{3*2=6.67} \;;\; 3*10=6 \;;\; 5*15=9 \;;\; .67+9=21 \;;\; 23-25=8 \;;\; 23-37=36 \;;\; 2*2=8 \;;\; 4+3=9 \;;\; 55*2=10 \;;\; 50/0.75=25 \;;\; 75.50+126=292.5 \;;\; 18*162=111 \;;\; 11-18=5 \;;\; 5*7=106 \;;\; 18*162=108 \;;\; 162-108=90 \;;\; 2+20=40 \;;\; 40+65=155 \;;\; 155/3=51 \;;\; 25*100=25 \;;\; 33/3=23 \;;\; 3/220=20.33 \;;\; 40+85=225 \;;\; 195-160=40 \;;\; 2*8=26 \;;\; 26/60=1 \;;\; 3*3=1 \;;\; 3+6=15 \;;\; 3*6=2 \;;\; 3*3=2 \;;\; 3*6=4 \;;\; 1/3=2 \;;\; 2/3=3.3 \;;\; .25+1.25=2.75 \;;\; 0-100=1700 \;;\; 400*3=120 \;;\; 1200-1000=2000 \;;\; 40-200=40 \;;\; 400-10=290.00 \;;\; 7+14=28 \;;\; 5+7=19 \;;\; 80+23=203 \;;\; 203-23=80 \;;\; 40/2=100 \;;\; 200*40=80 \;;\; 12/4=24 \;;\; 15*15=120 \;;\; 4*120=90 \;;\; \textbf{4-1=33} \;;\; 6*6=48 \;;\; 18+12=25 \;;\; 5000-3755=125 \;;\; 3750-5000=3750 \;;\; 25*40=10 \;;\; 100*25=5 \;;\; 4+4=9 \;;\; 2525-9=16 \;;\; 500+800=130 \;;\; 1300+13=139 \;;\; 1300+130=123 \;;\; 000+25000=1 \;;\; 126*60=756 \;;\; 100*840=84 \;;\; 840+756=149 \;;\; 140+280=605 \;;\; 140+340=625 \;;\; 5*88=40 \;;\; 2/100=15 \;;\; 13.35+34.2=84 \;;\; 128+256=512 \;;\; 16/2=1 \;;\; 1/2=8 \;;\; 8+4=33 \;;\; 4/3=8 \;;\; 4+2=30 \;;\; 740/110=6 \;;\; 180/110=16 \;;\; 4/4=100100 \;;\; 2*100=50 \;;\; 4+4=4 \;;\; 24/66=4 \;;\; 9+28=77 \;;\; 125-2=510 \;;\; 4-2=125 \;;\; 23*23=92 \;;\; 10/100=60 \;;\; 10/100=66 \;;\; 10/100=600 \;;\; 600600+600=6600 \;;\; 10/100=1320 \;;\; 15400+6000=22000 \;;\; 8800/88=1080 \;;\; 75*4=3 \;;\; 0*.110=22 \;;\;
}

A lot of these mistakes, although they seem far from correct can be easily explained in the context of MTP.  \\
\textbf{3*2=6.67} has an extra prediction of ".67`` due to 2-token prediction. The following CoT trace is conditioned on this wrong output. \\
\textbf{12/21=1} When 2-tokens are aggregated, both the numbers turned out the same, hence the result was 1. \\
\textbf{3*3=99} has an extra prediction of "9`` due to 2-token prediction. \\
and so on...

\section{C-MTP Ablation Studies}

In this section we present two ablation studies of C-MTP. The first study compares \mttrain{} with \wsinit{}, and the second ablates the necessity of $L_{KD}$ by training without distillation loss.

\subsection{\mttrain{} vs \wsinit{}}
\label{app:multitask}

We experiment with \mttrain{} and \wsinit{} on the full structured CoT trace dataset. Since C-MTP is a small addition on top of CoT, we view \wsinit{} as the more principled way but a fair comparison with CODI compelled us to experiment with the \mttrain{} setup. Our results show similar performance with both settings, hence we use \wsinit{} for our main table.

\begin{table*}[!t]
    \centering
    \setlength{\tabcolsep}{2pt}
    \caption{\textbf{Comparison of \wsinit{} and \mttrain{}}. with LLaMA-3.2-1B-Instruct on GSM8k-Aug (Structured) CoT traces.}
    \small
    \begin{tabular}{l cc cc cc cc}
        \toprule
        Method & \multicolumn{2}{c}{GSM8k-Test} & \multicolumn{2}{c}{GSM-Hard} & \multicolumn{2}{c}{SVAMP} & \multicolumn{2}{c}{MultiArith} \\
        \cmidrule(lr){2-3} \cmidrule(lr){4-5} \cmidrule(lr){6-7} \cmidrule(lr){8-9}
        & Acc. ($\uparrow$) & \# L ($\downarrow$) & Acc. ($\uparrow$) & \# L ($\downarrow$) & Acc. ($\uparrow$) & \# L ($\downarrow$) & Acc. ($\uparrow$) & \# L ($\downarrow$) \\
        \midrule
        \rowcolor{gray!15} CoT & 61.94 & $27.17_{\pm 11.24}$ & 14.22 & $34.43_{\pm 15.11}$ & 64.96 & $13.15_{\pm 5.30}$ & 98.89 & $15.46_{\pm 1.96}$ \\
        \midrule
        CODI & 53.03 & 6.00 & 12.40 & 6.00 & 60.26 & 6.00 & 99.11 & 6.00 \\
        CoLaR-3 & 32.65 & $9.01_{\pm 3.75}$ & 6.79 & $10.96_{\pm 6.30}$ & 48.36 & $4.49_{\pm 1.91}$ & 87.33 & $5.33_{\pm 1.93}$ \\
        \midrule
        \shortstack[l]{C-MTP-3 \\ (\wsinit{})} & 46.50 & $7.94_{\pm 3.69}$ & 10.58 & $9.02_{\pm 4.54}$ & 56.08 & $3.43_{\pm 1.66}$ & 97.33 & $4.16_{\pm 0.67}$ \\
        \midrule
        \shortstack[l]{C-MTP-3 \\ (\mttrain{})} & 48.29 & $7.81_{\pm 3.56}$ & 11.11 & $8.86_{\pm 4.05}$ & 55.02 & $3.44_{\pm 1.68}$ & 96.11 & $4.10_{\pm 0.58}$ \\
        \bottomrule
    \end{tabular}
\end{table*}

\subsection{Ablation on C-MTP: Training without $L_{KD}$}
\label{app:nodistill}

In this section, we ablate the effect of using the knowledge distillation loss $L_{KD}$ in our training setup. We compare the performance of our model with and without the $L_{KD}$ term, while keeping all other training configurations constant. The results are summarized in \cref{tab:trainaugllama_nokd} (Appendix). We observe that the performance difference is minimal, suggesting that the C-MTP training is able to learn effectively even without distilling from a frozen teacher. \wsinit{} from the CoT checkpoint provides a strong starting point and the simplicity of the C-MTP training objective prevents catastrophic forgetting.

\begin{table*}[h]
    \caption{\textbf{GSM8k-Aug results on LLaMA-3.2-1B-Instruct with no $L_{KD}$ on C-MTP.} Accuracy is averaged over 5 runs (run-to-run deviation $<$1\%). \#L denotes the mean chain-of-thought length (in steps) with standard deviation, computed over the full evaluation set.}
    \centering
    \setlength{\tabcolsep}{2pt}
    \small
    \begin{tabular}{ll cc cc cc cc}
        \toprule
        Training Method & & \multicolumn{2}{c}{GSM8k-Test} & \multicolumn{2}{c}{GSM-Hard} & \multicolumn{2}{c}{SVAMP} & \multicolumn{2}{c}{MultiArith} \\
        \cmidrule(lr){3-4} \cmidrule(lr){5-6} \cmidrule(lr){7-8} \cmidrule(lr){9-10}
        & & Acc. ($\uparrow$) & \# L ($\downarrow$) & Acc. ($\uparrow$) & \# L ($\downarrow$) & Acc. ($\uparrow$) & \# L ($\downarrow$) & Acc. ($\uparrow$) & \# L ($\downarrow$) \\
        \midrule
        \rowcolor{gray!15} Standard & CoT & 61.94 & $27.17_{\pm 11.24}$ & 14.22 & $34.43_{\pm 15.11}$ & 64.96 & $13.15_{\pm 5.30}$ & 98.89 & $15.46_{\pm 1.96}$ \\
        \midrule
        Indirect & CODI & 53.03 & 6.00 & 12.40 & 6.00 & 60.26 & 6.00 & \textbf{99.11} & 6.00 \\
        Supervision & Sim-CoT & \textbf{53.95} & 6.00 & 12.45 & \textbf{6.00} & 58.62 & 6.00 & 97.78 & 6.00 \\
        \midrule
        Direct & CoLaR-2 & 41.23 & $12.99_{\pm 5.53}$ & 8.87 & $16.30_{\pm 8.30}$ & 52.46 & $6.31_{\pm 2.66}$ & 90.56 & $7.39_{\pm 1.43}$ \\
        Supervision & CoLaR-3 & 32.65 & $9.01_{\pm 3.75}$ & 6.79 & $10.96_{\pm 6.30}$ & 48.36 & $4.49_{\pm 1.91}$ & 87.33 & $5.33_{\pm 1.93}$ \\
        & CoLaR-4 & 29.60 & $6.80_{\pm 2.71}$ & 6.60 & $7.66_{\pm 2.97}$ & 46.82 & $3.47_{\pm 1.48}$ & 88.11 & $4.13_{\pm 0.67}$ \\
        \cmidrule(lr){2-10}
        & C-MTP-2 & 53.30 & $12.07_{\pm 5.53}$ & \textbf{12.59} & $14.15_{\pm 6.44}$ & \textbf{62.10} & $5.14_{\pm 2.58}$ & 95.33 & $6.29_{\pm 1.20}$ \\
        & \quad-$L_{KD}$ & 52.12 & $12.10_{\pm 5.88}$ & 12.19 & $14.39_{\pm 6.94}$ & 59.50 & $5.27_{\pm 2.65}$ & 94.67 & $6.23_{\pm 0.98}$ \\
        & C-MTP-3 & 46.50 & $7.94_{\pm 3.69}$ & 10.58 & $9.02_{\pm 4.54}$ & 56.08 & $3.43_{\pm 1.66}$ & 97.33 & $4.16_{\pm 0.67}$ \\
        & \quad-$L_{KD}$ & 46.00 & $7.92_{\pm 3.65}$ & 10.60 & $8.99_{\pm 4.34}$ & 56.10 & $3.43_{\pm 1.69}$ & 96.67 & $4.17_{\pm 0.63}$ \\
        & C-MTP-4 & 40.70 & \bm{$5.84_{\pm 2.71}$} & 9.29 & $6.85_{\pm 4.34}$ & 50.98 & \bm{$2.39_{\pm 1.47}$} & 94.67 & \bm{$3.12_{\pm 0.56}$} \\
        & \quad-$L_{KD}$ & 36.60 & $5.84_{\pm 2.66}$ & 8.79 & $6.82_{\pm 3.22}$ & 48.70 & $2.42_{\pm 1.51}$ & 94.78 & $3.13_{\pm 0.51}$ \\
        \bottomrule
    \end{tabular}
    \label{tab:trainaugllama_nokd}
\end{table*}

\section{Extended Experimental Results}
\label{app:expdetails}
In this section we present additional details and results for the experiments described in the main paper.

\subsection{Setup Details}
\label{app:setup}

\Cref{tab:hyperparams} (Appendix) reports the training configuration for every run, following these conventions. \emph{Data} denotes the training corpus: \emph{Structured} (GSM8k-Aug, \texttt{icot} in code), \emph{Semi-Nat.} (semi-natural language CoT, \texttt{icot-nl} in code), \emph{GSM-Real.} (GSM-realistic, \texttt{gsmllama} in code), and \emph{Realistic} (MATH, \texttt{mathllama}/\texttt{mathqwen} in code). The effective batch size is reported as (per-device batch $\times$ gradient-accumulation steps). All runs use a warmup ratio of $0.05$, except the CODI runs, which use $0.03$ following the original implementation. All models are trained with LoRA adapters unless marked \emph{Full} (full finetuning). The last column reports the C-MTP span length or the number of CODI latent tokens, where applicable; it is left blank otherwise.

\begin{table*}[t]
    \caption{\textbf{Final hyperparameters for all experiments.} Learning rates use the shorthand \mbox{$5\text{e-}4 = 5\times10^{-4}$}. The effective batch size is reported as (per-device batch $\times$ gradient-accumulation steps).}
    \label{tab:hyperparams}
    \centering
    \small
    \setlength{\tabcolsep}{5pt}
    \begin{tabular}{lll cc c c c c}
        \toprule
        Backbone & Data & Setting & Batch (p$\times$g) & Ep. & LR & WD & Max Tok. & Span/Lat. \\
        \midrule
        \multicolumn{9}{l}{\textit{Explicit CoT}} \\
        LLaMA & Structured & LoRA & 128 (64$\times$2) & 4 & 5e-4 & 0.01 & 200 & -- \\
        LLaMA & Structured & Full & 128 (64$\times$2) & 3 & 2e-5 & 0.01 & 200 & -- \\
        LLaMA & Semi-Nat.  & LoRA & 128 (32$\times$4) & 5 & 5e-4 & 0.01 & 300 & -- \\
        LLaMA & Semi-Nat.  & Full & 128 (32$\times$4) & 4 & 2e-5 & 0.01 & 300 & -- \\
        Qwen  & Structured & LoRA & 128 (32$\times$4) & 5 & 5e-4 & 0.1  & 200 & -- \\
        Qwen  & Structured & Full & 128 (32$\times$4) & 3 & 2e-4 & 0.1  & 200 & -- \\
        LLaMA & GSM-Real.  & LoRA & 64 (32$\times$2)  & 10 & 5e-4 & 0.1  & 768 & -- \\
        LLaMA & Realistic  & LoRA & 64 (32$\times$2)  & 5 & 5e-4 & 0.1  & 1024 & -- \\
        Qwen  & Realistic  & LoRA & 64 (16$\times$4)  & 5 & 5e-4 & 0.1  & 1024 & -- \\
        \midrule
        \multicolumn{9}{l}{\textit{C-MTP}} \\
        LLaMA & Structured & LoRA          & 128 (32$\times$4) & 5 & 5e-4 & 0.01 & 200 & 2 / 3 / 4 \\
        LLaMA & Structured & LoRA (\mttrain{}) & 128 (32$\times$4) & 8 & 5e-4 & 0.01 & 200 & 3 \\
        LLaMA & Semi-Nat.  & LoRA          & 128 (32$\times$4) & 5 & 5e-4 & 0.01 & 300 & 2 / 3 / 4 \\
        Qwen  & Structured & LoRA          & 128 (32$\times$4) & 5 & 5e-4 & 0.1  & 200 & 2 / 3 / 4 \\
        LLaMA & GSM-Real.  & LoRA$^\dagger$ & 64 (8$\times$8)  & 5 & 5e-4 & 0.1  & 768 & 2 / 3 / 4 \\
        LLaMA & Realistic  & LoRA$^\dagger$ & 64 (8$\times$8)  & 8 & 5e-4 & 0.1  & 1024 & 2 \\
        Qwen  & Realistic  & LoRA$^\dagger$ & 64 (8$\times$8)  & 8 & 5e-4 & 0.1  & 1024 & 2 \\
        \midrule
        \multicolumn{9}{l}{\textit{CODI}} \\
        LLaMA & Structured & LoRA & 128 (64$\times$2) & 8  & 7e-4 & 0.1 & 200  & 6 \\
        LLaMA & Semi-Nat.  & LoRA & 128 (64$\times$2) & 8  & 7e-4 & 0.1 & 300  & 6 \\
        LLaMA & GSM-Real.  & LoRA & 128 (16$\times$8) & 4 / 5 / 5 & 5e-4 & 0.1 & 768 & 6 / 10 / 20 \\
        LLaMA & Realistic  & LoRA & 128 (16$\times$8) & 10 & 5e-4 & 0.1 & 1024 & 6 / 10 / 20 \\
        Qwen  & Realistic  & LoRA & 128 (16$\times$8) & 10 & 5e-4 & 0.1 & 1024 & 6 / 10 / 20 \\
        \bottomrule
    \end{tabular}

    \vspace{3pt}
    {\footnotesize $^\dagger$\,The C-MTP student and teacher are initialized from the merged LoRA checkpoint of the corresponding \emph{Realistic} CoT run (rows above).}
\end{table*}

All experiments were conducted on a single NVIDIA H200 GPU.

\subsection{Training on Structured CoT: GSM8k-Aug}
\label{app:structuredqwen}

We present results on Qwen2.5-1.5B-Instruct in \cref{tab:trainaugqwen} (Appendix). Our experiments on training Sim-CoT with this backbone were rife with training instabilities leading to suboptimal downstream results. Therefore, we did not add them in the table.

\begin{table*}[b]
    \caption{\textbf{GSM8k-Aug results on Qwen2.5-1.5B-Instruct.} Accuracy is averaged over 5 runs (run-to-run deviation $<$1\%). \#L denotes the mean chain-of-thought length (in steps) with standard deviation, computed over the full evaluation set.}
    \centering
    \setlength{\tabcolsep}{2pt}
    \small
    \begin{tabular}{ll cc cc cc cc}
        \toprule
        Training Method & & \multicolumn{2}{c}{GSM8k-Test} & \multicolumn{2}{c}{GSM-Hard} & \multicolumn{2}{c}{SVAMP} & \multicolumn{2}{c}{MultiArith} \\
        \cmidrule(lr){3-4} \cmidrule(lr){5-6} \cmidrule(lr){7-8} \cmidrule(lr){9-10}
        & & Acc. ($\uparrow$) & \# L ($\downarrow$) & Acc. ($\uparrow$) & \# L ($\downarrow$) & Acc. ($\uparrow$) & \# L ($\downarrow$) & Acc. ($\uparrow$) & \# L ($\downarrow$) \\
        \midrule
        \rowcolor{gray!15} Standard & CoT & 68.23 & $36.46_{\pm 16.56}$ & 26.23 & $57.37_{\pm 29.60}$ & 68.36 & $18.30_{\pm 9.24}$ & 98.00 & $19.18_{\pm 2.45}$ \\
        \midrule
        Indirect & CODI & 44.50 & $\bm{6.00}$ & 10.46 & $\bm{6.00}$ & \textbf{63.10} & $6.00$ & \textbf{92.22} & $6.00$ \\
        \midrule
        Direct & CoLaR-2 & 37.48 & $17.86_{\pm 8.40}$ & 9.01 & $27.55_{\pm 14.22}$ & 49.90 & $8.36_{\pm 4.38}$ & 84.33 & $9.30_{\pm 2.01}$ \\
        & CoLaR-3 & 30.69 & $12.04_{\pm 5.42}$ & 7.19 & $18.36_{\pm 10.05}$ & 43.64 & $5.80_{\pm 2.89}$ & 68.44 & $6.35_{\pm 1.03}$ \\
        & CoLaR-4 & 25.28 & $9.29_{\pm 4.13}$ & 6.02 & $14.14_{\pm 8.46}$ & 40.22 & $4.57_{\pm 2.17}$ & 73.33 & $4.88_{\pm 0.87}$ \\
        \cmidrule(lr){2-10}
        & C-MTP-2 & \textbf{50.57} & $16.60_{\pm 8.20}$ & \textbf{11.78} & $25.08_{\pm 12.56}$ & 54.20 & $7.51_{\pm 4.42}$ & 90.44 & $8.27_{\pm 1.47}$ \\
        & C-MTP-3 & 42.29 & $10.84_{\pm 5.52}$ & 10.60 & $16.01_{\pm 8.29}$ & 45.96 & $4.79_{\pm 2.88}$ & 79.22 & $5.53_{\pm 1.31}$ \\
        & C-MTP-4 & 34.42 & $7.99_{\pm 4.03}$ & 8.02 & $11.77_{\pm 6.29}$ & 44.24 & $\bm{3.40_{\pm 2.08}}$ & 72.11 & $\bm{3.94_{\pm 0.95}}$ \\
        \bottomrule
    \end{tabular}
    \label{tab:trainaugqwen}
\end{table*}

\subsection{Sample Efficiency}
\label{app:sampeff}

\begin{figure}[h]
    \centering
    \caption{Sample efficiency comparison on GSM8k-Aug.}
    \includegraphics[width=\textwidth]{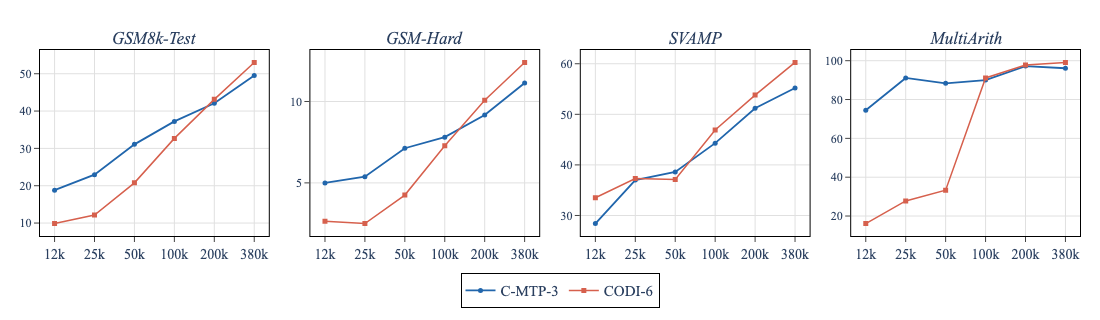}
    \label{fig:sampleff}
\end{figure}

To understand how the two training paradigms compare under varying amounts of supervision, we train both methods on randomly sampled subsets of GSM8k-Aug (12k, 25k, 50k, 100k and 200k samples) and evaluate on downstream tasks. For fair comparison with CODI, we use the \mttrain{} configuration where we train the same backbone for CoT and C-MTP and use the CoT inference output as the teacher. For each random subset size we run a grid search over $lr \in$ \{5e-4, 7e-4, 8e-4\} and $epochs \in$ \{6, 7, 10\} and select the configuration with the final best validation loss. \cref{fig:sampleff} (Appendix) shows accuracy as a function of training set size. In the low-data regime (up to $\sim$100k samples), C-MTP consistently outperforms CODI, with the gap most pronounced at 12k samples. On in-distribution tasks (GSM8k-Test, GSM-Hard), C-MTP's advantage is particularly clear, whereas on out-of-distribution tasks (SVAMP, MultiArith) the gap is smaller and more variable.

\subsection{Fixed Wallclock Time Comparison}
\label{app:fixwallclock}

Based on our analysis (\cref{tab:timing_analysis}, Appendix), we estimate that a single C-MTP training step is on average $2\times$ faster than CODI on LLaMA-3.2-1B-Instruct, allowing C-MTP to process roughly twice as many samples within the same wallclock budget.

In a finetuning setup, this increased compute can be allocated in several ways, such as additional samples, more epochs, a larger base model, or increased sequence length. Here, we isolate the effect of increased sample count under direct supervision (C-MTP) versus indirect supervision (CODI). To maintain wallclock parity under this estimated $2\times$ factor, we hold the number of epochs constant and set learning rates based on sample-efficiency grid sweeps. We note that this configuration is not necessarily optimal and serves primarily to illustrate the gains achievable under a fixed time budget. Overall, \cref{tab:fixedwallclock} (Appendix) shows that C-MTP-4 consistently outperforms CODI across all budget levels, with particularly pronounced gains on GSM8k-Test.

\begin{table}[t]
    \caption{\rev{\change{9} Estimated Fixed Training Wallclock Time Comparison on GSM8k-Aug}. We report downstream accuracy on each (\# Samples, Epochs) configuration on CODI and C-MTP-4. C-MTP-4 consistently outperforms CODI across all budget levels, with gains on GSM8k-Test and MultiArith particularly pronounced.}
    \centering
    \setlength{\tabcolsep}{2pt}
    \small
    \begin{tabular}{cl cccc}
        \toprule
        \shortstack{Wallclock \\ Time} & & \shortstack{GSM8k \\ Test} & \shortstack{GSM \\ Hard} & \shortstack{SVAMP \\ \vspace{0.1em}} & \shortstack{MultiArith \\ \vspace{0.1em}} \\
        \midrule
        (50k, 7) & CODI & 20.85 & 4.17 & 36.80 & 33.33 \\
        \rowcolor{gray!15} (100k, 7) & C-MTP-4 & 29.19 & 6.29 & 43.10 & 86.11 \\
        \midrule
        (60k, 6) & CODI & 22.44 & 5.00 & 40.90 & 31.67 \\
        \rowcolor{gray!15} (120k, 6) & C-MTP-4 & 30.63 & 6.82 & 41.90 & 87.22 \\
        \midrule
        (70k, 5) & CODI & 22.14 & 4.62 & 42.70 & 37.22 \\
        \rowcolor{gray!15} (140k, 5) & C-MTP-4 & 32.37 & 7.96 & 41.90 & 86.67 \\
        \midrule
        (88k, 4) & CODI & 26.00 & 5.46 & 42.90 & 51.11 \\
        \rowcolor{gray!15} (175k, 4) & C-MTP-4 & 33.59 & 8.42 & 45.20 & 91.67 \\
        \bottomrule
    \end{tabular}
    \label{tab:fixedwallclock}
\end{table}

\subsection{Generalization to Problem Structures}
\label{app:genproblems}

GSM8k-Aug \citep{DBLP:journals/corr/abs-2405-14838} was constructed by prompting GPT-4 to generate problems similar to GSM8k, thereby expanding the dataset $\sim50\times$ via prompting and heuristic filtering. This process results in  substantial structural redundancy: our analysis shows that over 50\% of all training samples reduce to only 100 distinct problem structures. To evaluate out-of-domain generalization to problem structures not seen during training, we partition the data by held-out problem structures. More analysis in \cref{app:gsmauganal}.

Concretely, let $\mathcal{D}^n \subset \mathcal{D}$ denote the subset of
samples containing $n$ reasoning steps, for $n \in \{2,3,4\}$. We partition
each $\mathcal{D}^n$ by problem structure into two disjoint, roughly
equal-sized splits $\mathcal{D}^n_A$ and $\mathcal{D}^n_B$, ensuring no
structure appears in both. Within each split, we hold out 2\% of problems
as a test set. For each $(n, X) \in \{2,3,4\} \times \{A, B\}$, we train
on $\mathcal{D}^n_X$ and evaluate on the test sets of \emph{both}
$\mathcal{D}^n_A$ and $\mathcal{D}^n_B$, yielding in-domain accuracy
(same split) and out-of-domain accuracy (held-out structures).

To ensure a fair comparison, we first train C-MTP-4 on all splits and measure the average generation length on each test set: $\mathcal{D}^2{=}3.23 \pm 0.55$, $\mathcal{D}^3{=}5.38 \pm 0.75$, $\mathcal{D}^4{=}7.04 \pm 1.13$ tokens. Since CODI requires the number of latents to be specified before training, we use these lengths to guide the choice, setting the latent count to 4, 6, and 8 for $n \in \{2, 3, 4\}$ respectively.

\cref{tab:gridsall} (Appendix) reports in-domain and out-of-domain accuracy for C-MTP and CODI across reasoning lengths $n \in \{2, 3, 4\}$. C-MTP generalizes better to held-out problem structures in most settings. The gap is particularly clear at $n=3$ and $n=4$, where CODI's out-of-domain accuracy drops substantially (e.g., 27.10 and 13.83 on $n=3$ and $n=4$ respectively) while C-MTP retains stronger performance (42.86 and 24.04). In-domain, the two methods are more competitive, with CODI performing slightly better at $n=2$ while C-MTP leads at longer reasoning chains. Overall, these results suggest that the sequential training of indirect supervision methods leads to overfitting to the specific problem structures seen during training.

{
\renewcommand{\arraystretch}{1.5}
\begin{table}[t]
\caption{\colorbox{indomain}{in-domain} and \colorbox{outdomain}{out-of-domain} evaluation results for C-MTP-4 and CODI across reasoning lengths $n \in \{2, 3, 4\}$. Each matrix entry reports accuracy when trained on the row set and evaluated on the column set.}
\centering
\begin{minipage}{0.3\columnwidth}
\centering
\textbf{\large C-MTP}
\end{minipage}%
\begin{minipage}{0.3\columnwidth}
\centering
\textbf{\large CODI}
\end{minipage}

\begin{minipage}{0.3\columnwidth}
\centering
\begin{tabular}{ccc}
                  & $\mathcal{D}^2_A$ & $\mathcal{D}^2_B$ \\
                  \cmidrule{2-3}
$\mathcal{D}^2_A$ & \cellcolor{indomain}$73.41$  & \cellcolor{outdomain}\underline{$53.54$} \\
$\mathcal{D}^2_B$ & \cellcolor{outdomain}$60.40$ & \cellcolor{indomain}$72.42$  \\
\end{tabular}
\end{minipage}%
\begin{minipage}{0.3\columnwidth}
\centering
\begin{tabular}{ccc}
                  & $\mathcal{D}^2_A$ & $\mathcal{D}^2_B$ \\
                  \cmidrule{2-3}
$\mathcal{D}^2_A$ & \cellcolor{indomain}\underline{$75.11$}  & \cellcolor{outdomain}$51.56$ \\
$\mathcal{D}^2_B$ & \cellcolor{outdomain}\underline{$63.79$} & \cellcolor{indomain}\underline{$76.52$}  \\
\end{tabular}
\end{minipage}%

\vspace{1.5em}
\begin{minipage}{0.3\columnwidth}
\centering
\begin{tabular}{ccc}
                  & $\mathcal{D}^3_A$ & $\mathcal{D}^3_B$ \\
                  \cmidrule{2-3}
$\mathcal{D}^3_A$ & \cellcolor{indomain}\underline{$56.46$}  & \cellcolor{outdomain}\underline{$42.86$} \\
$\mathcal{D}^3_B$ & \cellcolor{outdomain}\underline{$46.09$} & \cellcolor{indomain}\underline{$55.38$}  \\
\end{tabular}
\end{minipage}%
\begin{minipage}{0.3\columnwidth}
\centering
\begin{tabular}{ccc}
                  & $\mathcal{D}^3_A$ & $\mathcal{D}^3_B$ \\
                  \cmidrule{2-3}
$\mathcal{D}^3_A$ & \cellcolor{indomain}$41.88$  & \cellcolor{outdomain}$27.10$ \\
$\mathcal{D}^3_B$ & \cellcolor{outdomain}$26.42$ & \cellcolor{indomain}$40.41$  \\
\end{tabular}
\end{minipage}%

\vspace{1.5em}
\begin{minipage}{0.3\columnwidth}
\centering
\begin{tabular}{ccc}
                  & $\mathcal{D}^4_A$ & $\mathcal{D}^4_B$ \\
                  \cmidrule{2-3}
$\mathcal{D}^4_A$ & \cellcolor{indomain}\underline{$32.55$}  & \cellcolor{outdomain}\underline{$24.04$} \\
$\mathcal{D}^4_B$ & \cellcolor{outdomain}\underline{$24.68$} & \cellcolor{indomain}\underline{$33.40$}  \\
\end{tabular}
\end{minipage}%
\begin{minipage}{0.3\columnwidth}
\centering
\begin{tabular}{ccc}
                  & $\mathcal{D}^4_A$ & $\mathcal{D}^4_B$ \\
                  \cmidrule{2-3}
$\mathcal{D}^4_A$ & \cellcolor{indomain}$17.87$  & \cellcolor{outdomain}$13.83$ \\
$\mathcal{D}^4_B$ & \cellcolor{outdomain}$17.23$ & \cellcolor{indomain}$18.51$  \\
\end{tabular}
\end{minipage}
\vspace{1em}
\label{tab:gridsall}
\end{table}
}

\subsection{Gold Trace Analysis}
\label{app:goldtrace}

\cref{tab:goldtrace} reports the numerical accuracies used to plot figures in \cref{fig:error_and_goldfill}. We also include the result on SVAMP with structured (GSM8k-Aug) traces. As we add in more of the reasoning trace the accuracy improves, reaching almost CoT accuracy with full reasoning trace, showing that aggregation is not the main cause of regression but the hardness of predicting next latents in the aggregated space. For structured traces, no ground truth exists, so we use correct CoT generations to perform sweeps, resulting in CoT accuracy being $100\%$.

\begin{table}[h]
    \caption{\textbf{Gold-prefill sweep} Accuracy (\%) as a function of the prefilled fraction $x$ of the gold trace. $x=0$ is the original evaluation (greedy decoding) and $x=1$ feeds the full gold trace to produce answer directly.}
    \label{tab:goldtrace}
    \centering
    \small
    \setlength{\tabcolsep}{3.5pt}
    \begin{tabular}{c cccc ccc ccc}
        \toprule
        & \multicolumn{4}{c}{GSM8k-Aug-NL, GSM8k-test} & \multicolumn{3}{c}{GSM8k-Aug, GSM8k-test} & \multicolumn{3}{c}{GSM8k-Aug, SVAMP} \\
        \cmidrule(lr){2-5} \cmidrule(lr){6-8} \cmidrule(lr){9-11}
        $x$ & CoT & C-MTP-2 & C-MTP-3 & C-MTP-4 & C-MTP-2 & C-MTP-3 & C-MTP-4 & C-MTP-2 & C-MTP-3 & C-MTP-4 \\
        \midrule
        0.0 & 53.07 & 44.58 & 35.86 & 26.08 & 75.88 & 67.36 & 59.81 & 84.39 & 77.59 & 71.72 \\
        0.1 & 57.54 & 45.64 & 36.62 & 28.28 & 76.74 & 69.43 & 62.12 & 84.39 & 77.74 & 71.56 \\
        0.2 & 57.70 & 42.00 & 30.25 & 23.28 & 78.56 & 72.35 & 65.41 & 87.64 & 86.55 & 74.96 \\
        0.3 & 59.67 & 42.08 & 29.80 & 21.23 & 80.51 & 74.30 & 66.26 & 88.87 & 87.02 & 84.08 \\
        0.4 & 62.47 & 43.97 & 30.63 & 23.05 & 82.83 & 76.49 & 69.67 & 93.20 & 87.48 & 84.85 \\
        0.5 & 65.88 & 49.05 & 35.33 & 26.54 & 85.99 & 80.51 & 75.27 & 93.66 & 90.88 & 86.24 \\
        0.6 & 70.13 & 51.93 & 39.27 & 32.37 & 89.28 & 84.90 & 78.20 & 93.82 & 92.89 & 86.71 \\
        0.7 & 77.48 & 59.67 & 42.76 & 38.21 & 92.33 & 89.65 & 84.65 & 95.05 & 94.28 & 88.87 \\
        0.8 & 84.46 & 66.11 & 52.54 & 49.66 & 94.28 & 91.72 & 91.60 & 95.36 & 95.05 & 97.37 \\
        0.9 & 92.80 & 71.49 & 62.93 & 63.61 & 97.08 & 96.95 & 95.86 & 98.61 & 98.61 & 99.85 \\
        \midrule
        1.0 & 99.09 & 94.39 & 90.98 & 87.79 & 99.88 & 97.44 & 99.51 & 100.00 & 99.23 & 99.85 \\
        \bottomrule
    \end{tabular}
\end{table}

\subsection{Ablation Studies}
\label{sec:ablations}

\begin{table}[h]
    \centering
    \captionof{table}{\textbf{C-MTP ablations on GSM8k-Aug (LLaMA-3.2-1B-Instruct).} Accuracy ($\uparrow$), averaged over 5 runs. Default C-MTP rows use \wsinit{} with $L_{KD}$; sub-rows ablate the distillation loss (w/o $L_{KD}$) and the training configuration (\mttrain{}). Full tables with reasoning lengths in \cref{app:multitask,app:nodistill}.}
    \label{tab:ablations}
    \setlength{\tabcolsep}{3pt}
    \small
    \begin{tabular}{l cccc}
        \toprule
        Method & \shortstack{GSM8k\\Test} & \shortstack{GSM\\Hard} & SVAMP & \shortstack{Multi\\Arith} \\
        \midrule
        \rowcolor{gray!15} CoT & 61.94 & 14.22 & 64.96 & 98.89 \\
        \midrule
        C-MTP-2            & 53.30 & 12.59 & 62.10 & 95.33 \\
        \quad w/o $L_{KD}$ & 52.12 & 12.19 & 59.50 & 94.67 \\
        \midrule
        C-MTP-3            & 46.50 & 10.58 & 56.08 & 97.33 \\
        \quad w/o $L_{KD}$ & 46.00 & 10.60 & 56.10 & 96.67 \\
        \quad \mttrain{} & 48.29 & 11.11 & 55.02 & 96.11 \\
        \midrule
        C-MTP-4            & 40.70 & 9.29  & 50.98 & 94.67 \\
        \quad w/o $L_{KD}$ & 36.60 & 8.79  & 48.70 & 94.78 \\
        \bottomrule
    \end{tabular}
\end{table}

\newpage

\subsection{Full Example of MTP Error Accumulation}
\label{app:mtperrors}

We present an example of how an error at the initial part of the sequence can cascade to the rest of the CoT. / represents span prediction boundary.

\small {
\begin{tcolorbox}[width=\textwidth]

\textbf{Question:} Janet has 16 eggs. She uses 3 eggs to make breakfast and 4 eggs to bake muffins. If she sells the remaining eggs at the farmers' market for 2 each, how much money does she make at the farmers' market each day? \\
\textbf{CoT:} She has 16 eggs - 3 eggs - 4 eggs = 9 eggs left for the farmers' market. Thus, Janet makes 9 * $2 = $18 in the farmers' market each day. \\
\textbf{MTP Prediction:}Janet/ has /16 -/ 3/ = /13 eggs/ left after/ her breakfast/. She/ bakes/ muffins/ with /13 \makered{*}/ 4/ = /\makered{52} eggs/. She/ sells /\makered{52} -/ 16/ = /36 eggs/ at the/ farmers'/ market daily/. So/, she/ makes /36 */ 2/ = /72 daily/ at the/ farmers'/ market.

\end{tcolorbox}

}

\small {
\begin{tcolorbox}[width=\textwidth]

\textbf{Question:} Emma can make and upload 72 vlogs per month. But she was only able to make 18 vlogs for the first week, 21 vlogs for the second week, and 15 vlogs for the third week. How many vlogs should she do to complete the 72 vlogs per month? \\
\textbf{CoT:} Emma was able to make 18 + 21 + 15 = 54 vlogs for the first three weeks. So, she needs to make 72 - 54 = 18 more vlogs to meet the 72 vlogs per month requirement. \\
\textbf{MTP Prediction:} Emma has first/ to  /72 - /18 - /21 = /\makered{45} vib/logs to meet/ her quota./ Finally her third/ week, she/ has  /\makered{45} - /15 = /30 moreib/ toib to/ meet. quota/ Thus, of/ she should /72 + /30 = /102 moreks/ to meet the/ monthly72.

\end{tcolorbox}

}

\subsection{Experiments on CommonsenseQA}
\label{app:commonsenseexps}

We experiment with CommonsenseQA \citep{talmor-etal-2019-commonsenseqa}, a commonsense multiple-choice question-answering dataset paired with natural language CoTs generated by GPT-4o-mini \citep{DBLP:conf/emnlp/ShenYZHDH25}. We see that Indirect Supervision (CODI) remains preferable even exceeding CoT-SFT in this case; their compression objective allows them to discard the filler words and focus on the key associations required to get the final answer.
For direct supervision (C-MTP), we find that although the traces are nearly as long as semi-natural math traces, we do not see similar degradation with increasing span lengths. Natural language, devoid of mathematically precise tokens, makes the final answer robust to small errors in multi-token prediction.

\begin{table}[!h]
    \renewcommand{\arraystretch}{1.15}
    \caption{Results on CommonsenseQA with Qwen2.5-1.5B-Instruct. }
    \centering
    \small
    \begin{tabular}{ll cc}
        \toprule
        Training & & Acc. ($\uparrow$) & \# L ($\downarrow$) \\
        \midrule
        Standard & CoT-SFT  & $74.40$ & $79.21_{\pm 40.34}$ \\
        \midrule
        Indirect & CODI-6   & $77.12$ & 6.00 \\
        \midrule
        \multirow{6}{*}{Direct} & C-MTP-2  & $70.73$ & $44.37_{\pm 21.50}$ \\
        & C-MTP-3 & $70.52$ & $31.06_{\pm 14.22}$ \\
        & C-MTP-4 & $69.25$ & $22.90_{\pm 10.70}$ \\
        & C-MTP-5 & $70.96$ & $17.68_{\pm 8.15}$ \\
        \bottomrule
    \end{tabular}
    \label{tab:commonsenseqa}
\end{table}

\small {
\begin{tcolorbox}[width=\textwidth]

\textbf{Question:} Where would you put a glass after drinking from it?\texttt{\textbackslash n} Choices:\texttt{\textbackslash n} A: ocean\texttt{\textbackslash n} B: water cooler\texttt{\textbackslash n} C: cabinet\texttt{\textbackslash n} D: dishwasher\texttt{\textbackslash n} E: dining room \\
\vspace{0.7em}

\textbf{C-MTP-5 Thinking:} After finishing a bottle,/ the most logical place would/ be be a sink it/ in, it, it/ is it for the for/ cleaning.. The options/ other either do not,/ logically the to the./ other, is\;\;/ the water,,/ dishwasher, is the the/ most appropriate for for/ cleaning.
\vspace{0.7em}

\textbf{C-MTP-5 Final Answer:} The answer is: D \\
\textbf{Ground truth:} D

\end{tcolorbox}

}

\rev{\change{8}
\section{Details on Realistic CoT Datasets}
\label{app:realisticdetails}

For both \emph{realistic} tasks on LLaMA-3.2-1B-Instruct, we use its \href{https://huggingface.co/datasets/meta-llama/Llama-3.2-1B-Instruct-evals}{official evaluation prompt} without the in-context examples. For Qwen, we use the standard CoT prompt. Since these responses are closest in distribution to LLM generations, we hypothesize they should be easiest to learn. Consequently, \cref{tab:trainmath} shows minimal performance change upon CoT finetuning.

For both \emph{realistic} tasks, we generate a response to every training problem and keep cot traces only for the problems whose generated final answer is correct. The problem count therefore differs across settings (\cref{tab:tracelengths}). For GSM8k-semi-natural to realistic (with LLaMA-3.2-1B-Instruct), number of samples becomes $\approx 0.5x$ but the average length increases $\approx 2.5x$.

\begin{table}[h]
    \centering
    \caption{Training set size and trace length(in tokens) across settings for LLaMA-3.2-1B-Instruct. The GSM8k-Aug train split contains 380k problems, and the MATH train split contains 10,371 problems after removing problems whose solution exceeds 512 tokens. Realistic setting uses a subset of these problem sets.}
    \begin{tabular}{llcc}
        \toprule
        Setting & Problem Source & \# Train Problems & Trace Length (mean $\pm$ std) \\
        \midrule
        GSM structured & GSM8k-Aug & 380,015 & $21.46 \pm 10.46$ \\
        GSM semi-natural & GSM8k-Aug & 380,013 & $55.98 \pm 27.68$ \\
        GSM realistic & GSM8k-Aug (model-solved) & 206,701 & $134.29 \pm 42.50$ \\
        MATH realistic & MATH (model-solved) & 3,587 & $242.67 \pm 146.69$ \\
        \bottomrule
    \end{tabular}
    \label{tab:tracelengths}
\end{table}

\begin{table}[h]
    \centering
    \caption{Summary of accuracy across settings on LLaMA-3.2-1B-Instruct. GSM settings report the weighted average over GSM8k-Test, GSM-Hard, SVAMP, and MultiArith (weighted by test-set size); MATH reports accuracy on the MATH test set. Parentheses give the relative drop from CoT-SFT for each configuration.}
    \small
    \begin{tabular}{l cccc}
        \toprule
        & GSM structured & GSM semi-natural & GSM realistic & MATH realistic \\
        \midrule
        CoT-SFT & 47.99 & 43.54 & 37.82 & 23.80 \\
        \midrule
        C-MTP-2 & 43.52 (9.3\%) & 37.65 (13.5\%) & 31.18 (17.6\%) & 7.40 (68.9\%) \\
        CODI-6 & 43.06 (10.3\%) & 39.70 (8.8\%) & 24.37 (35.6\%) & 8.52 (64.2\%) \\
        CODI-10 & & & 25.97 (31.3\%) & 9.92 (58.3\%) \\
        \bottomrule
    \end{tabular}
    \label{tab:summarydrops}
\end{table}

Qwen2.5-1.5B-Instruct provides correct solutions to 5,976 problems on MATH (up from 3.5k for LLaMA), but we still notice the drop in performance, suggesting that the CoT-SFT and Latent reasoning gap might exist with more data. Experiments with larger datasets are a natural next step.

}

\section{CoLaR - CODI Implementation Discrepancy}

We acknowledge that our reproduced results for CODI \citep{DBLP:conf/emnlp/ShenYZHDH25} differ from the baseline results reported by CoLaR \citep{tan2026think}. This discrepancy likely stems from implementation differences, as the official CODI codebase may not have been accessible during CoLaR's development. Our experiments use the official CODI repository, yielding results consistent with those reported in the original paper.

\end{document}